\PassOptionsToPackage{small,bf}{caption}

\documentclass[smallextended]{svjour3}

\RequirePackage{fix-cm}
\usepackage{mathptmx} 
\smartqed
\journalname{Language Resources and Evaluation Journal}

\usepackage{url}
\usepackage{epsfig}
\usepackage{latexsym}
\usepackage{microtype}
\usepackage{tabularx}
\usepackage{amsmath}
\usepackage{arydshln}
\usepackage{booktabs}
\usepackage{wrapfig}
\usepackage{pgfplots}
\usepackage{textcomp}
\usepackage{chngpage}
\usepackage{graphicx,adjustbox}
\usepackage{multirow, makecell}
\usepackage{hyperref}
\usepackage[group-separator={,}, group-four-digits=true, group-digits=integer]{siunitx}
\usepackage[notext,not1]{stix}
\usepackage{fancyvrb}
\usepackage{natbib}
\usepackage{tabularx}
\usepackage{colortbl}
\usepackage{xcolor}
\usepackage[T5,T1]{fontenc}
\pgfplotsset{compat=1.7}

\usepackage{tikz}
\usepackage[linguistics]{forest}

\let\citewpar\citep
\let\citewopar\citet

\begin{document}
\title{DravidianCodeMix: Sentiment Analysis and Offensive Language Identification Dataset for Dravidian Languages in Code-Mixed Text
}
%


\titlerunning{DravidianCodeMix: Classification Dataset for Dravidian Languages Code-Mixed Text}        

\author{Bharathi Raja Chakravarthi\(^1\) \and
          Ruba Priyadharshini\(^2\)  \and
          Vigneshwaran Muralidaran\(^3\) \and 
          Navya Jose\(^4\) \and 
          Shardul Suryawanshi\(^1\) \and
          Elizabeth Sherly\(^4\) \and 
          John P. McCrae\(^1\)
}

\authorrunning{Chakravarthi et al} 

\institute{Bharathi Raja Chakravarthi* \\
              \email{bharathi.raja@insight-centre.org} \\          
         \\
         Ruba Priyadharshini \\
              \email{rubapriyadharshini.a@gmail.com} \\    
                \\
         Vigneshwaran Muralidaran \\
              \email{MuralidaranV@cardiff.ac.uk} \\
                                   \\
         Navya Jose \\
              \email{navya.mi3@iiitmk.ac.in} \\ 
                                     \\
         Shardul Suryawanshi \\
              \email{shardul.suryawanshi@insight-centre.org} \\ 
                       \\
         Elizabeth Sherly \\
              \email{sherly@iiitmk.ac.in} \\ 
                                     \\
         John P. McCrae \\
              \email{john.mccrae@insight-centre.org} \\ 
              \\
          \(^1\)Insight SFI Research Centre for Data Analytics, Data Science Institute, National University of Ireland Galway, Galway, Ireland \\
          \(^2\)ULTRA Arts and Science College, Madurai, Tamil Nadu, India \\
          \(^3\)School of Computer Science and Informatics, Cardiff University, Cardiff, United Kingdom \\
           \(^4\)Indian Institute of Information  Technology and Management-Kerala, Kerala, India
}

\date{Received: date / Accepted: date}
\maketitle
\begin{abstract}
This paper describes the development of a multilingual, manually annotated dataset for three under-resourced Dravidian languages generated from social media comments. The dataset was annotated for sentiment analysis and offensive language identification for a total of more than 60,000 YouTube comments. The dataset consists of around 44,000 comments in Tamil-English, around 7,000 comments in Kannada-English, and around 20,000 comments in Malayalam-English. The data was manually annotated by volunteer annotators and has a high inter-annotator agreement in Krippendorff's alpha. The dataset contains all types of code-mixing phenomena since it comprises user-generated content from a multilingual country.  We also present baseline experiments to establish benchmarks on the dataset using machine learning methods. The dataset is available on Github \footnote{\url{https://github.com/bharathichezhiyan/DravidianCodeMix-Dataset}} and Zenodo \footnote{ \url{https://zenodo.org/record/4750858\#.YJtw0SYo\_0M}}.
\end{abstract}
\section{Introduction}
Sentiment analysis is the classification task of mining sentiments from natural language, which finds use in numerous applications such as reputation management, customer support, and moderating content in social media \citewpar{wilson-etal-2005-recognizing,agarwal-etal-2011-sentiment,sajeetha9063341,sajeetha9185369}. Sentiment analysis has helped  industry to compile a summary of human perspectives and interests derived from feedback or even just the polarity of comments \citewpar{pang-lee-2004-sentimental,sajeetha9342640}.  Offensive language identification is another classification task in natural language processing (NLP), where the aim is to moderate and minimise offensive content in social media. In recent years, sentiment analysis and offensive language identification have gained significant interest in the field of NLP. 

Social media websites and product review forums provide opportunities for users to create content in informal settings. Moreover, to improve user experience, these platforms ensure that the user communicates his/her opinion in such a way that he/she feels comfortable either using native language or switching between one or more languages in the same conversation \citewpar{vyas-etal-2014-pos}. However, most NLP systems are trained on languages in formal settings with proper grammar,  which creates issues when it comes to the analysis phase of "user generated" comments \citewpar{chanda-etal-2016-unraveling,pratapa-etal-2018-language}.  Further, most of the developments in sentiment analysis and offensive language identification systems are performed on monolingual data for high-resource languages, while the user-generated content in under-resourced settings are often mixed with English or other high-resource languages \citewpar{winata-etal-2019-learning,chakravarthi-code-mix-survey}.

Code-mixing or code-switching is the alternation between two or more languages at the level of the document, paragraph, comments, sentence, phrase, word or morpheme. It is a distinctive aspect of conversation or dialogue in bilingual and multilingual societies \citewpar{barman-etal-2014-code}. It is motivated by structural, discourse, pragmatic and socio-linguistic reasons \citewpar{sridhar1978functions}. Most of the social media comments are code-mixed, while the resources created for sentiment analysis and offensive language identification are primarily available for monolingual texts. Code-mixing occurs in daily life, such as in normal conversation or social media conversation in both audio and text format. Code-mixing refers to the way a bilingual/ multilingual speaker changes his or her utterance into another language.  The vast majority of language pairs are under-resourced with regards to code-mixing tasks \citewpar{bali-etal-2014-borrowing,chakravarthi-code-mix-survey}. 

In this paper, we describe the creation of a corpus for Dravidian languages in the context of sentiment analysis and offensive language detection tasks. Dravidian languages are spoken mainly in the south of India \citewpar{chakravarthi-etal-2020-senti-tamil}. The four major literary languages belonging to the language family are Tamil (ISO 639-3: tam), Telugu (ISO 639-3: tel), Malayalam (ISO 639-3: mal), and Kannada (ISO 639-3: kan). Tamil, Malayalam and Kannada fall under the South Dravidian subgroup while Telugu belongs to the South Central Dravidian subgroup \citewpar{vikram2007development}. Each of the four languages has official status as one of the 22 scheduled languages recognised by the Government of India. Tamil also has official status in Sri Lanka and Singapore \citewpar{thamburaj2015critical}. Although the languages are widely spoken by millions of people, the tools and resources available for building robust NLP applications are under-developed for these languages.

Dravidian languages are highly agglutinating languages and each language uses its own script \citewpar{krishnamurti2003dravidian,sakuntharaj2016novel,sakuntharaj2017use}. The writing system is a phonemic abugida written from left to right. The Tamil language was written using   Tamili, Vattezhuthu, Chola, Pallava and Chola-Pallava scripts at different points in history. The modern Tamil script descended from the Chola-Pallava script that was conceived around the 4th century CE  \citewpar{sakuntharaj2018detecting,sakuntharaj2018refined}. The Malayalam script is based on the Vatteluttu script developed from old Vatteluttu with additional letters from  Grantha script to write loan words \citewpar{thottingal-2019-finite}. Similarly, the Kannada and Telugu scripts evolved from Bhattiprolu Brahmi. Nevertheless, social media users often use the Latin script for typing in these languages due to its ease of use and accessibility in handheld devices and computers \citewpar{thamburaj2015analysis}.

Monolingual datasets are available for Indian languages for various research aims \citewpar{agrawal-etal-2018-beating,thenmozhi2018ontology,kumar2020tamil}. However, there have been few attempts to make datasets for Tamil, Kannada and Malayalam code-mixed text \citewpar{chakravarthi-etal-2020-senti-malayalam,chakravarthi-etal-2020-senti-tamil,chakravarthi-2020-hopeedi,chakravarthi-muralidaran-2021-findings}. We believe it is essential to come up with approaches to tackle this resource bottleneck so that these languages can be equipped with NLP support in social media in a way that is both cost-effective and rapid. To create resources for a Tamil-English, Kannada-English and Malayalam-English code-mixed scenario, we collected comments on various Tamil, Kannada and Malayalam movie trailers from YouTube.

\begin{figure*}
  \includegraphics[width=\textwidth]{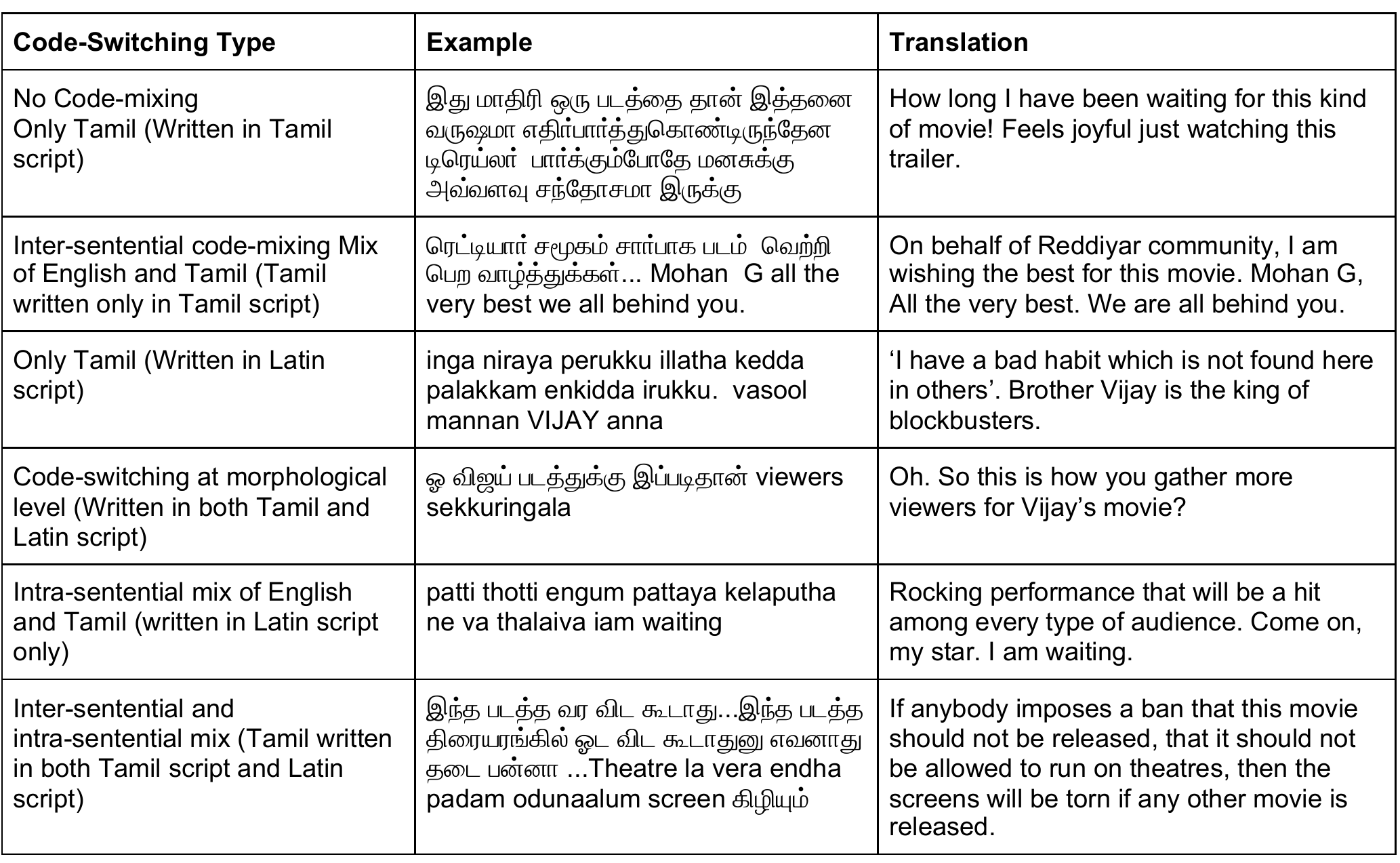}
  \caption{Examples of code mixing in Tamil dataset.}
  \label{fig:example-codemix-ta}
\end{figure*}

The contributions of this paper are:
\begin{enumerate}
  \item We present the dataset for three Dravidian languages, namely Tamil, Kannada, and Malayalam,  for sentiment analysis and offensive language identification tasks.
  \item The dataset contains all types\footnote{different types of code-mixing are shown in Figure \ref{fig:example-codemix-ta}} of code-mixing. This is the first Dravidian language dataset to contain all types of code-mixing, including mixtures of these scripts and the Latin script. The dataset consists of around 44,000 comments in Tamil-English, around 7,000 comments in Kannada-English, and around 20,000 comments in Malayalam-English.
\item We provide an experimental analysis of logistic regression, naive Bayes, decision tree, random forest, and SVM  on our code-mixed data for classification tasks so as to create a benchmark for further research.
\end{enumerate}
\section{Related Work} 
Sentiment analysis helps to understand the polarity (positive, negative or neutral) of the audience towards a content (comment, tweet, image, video) or an event (Brexit, presidential elections). This data on polarity can help in understanding public opinion. Furthermore, the inclusion of sentiment analysis can improve the performance of tasks such as recommendation system \citewpar{krishna2013learning,musto2017multi}, and hate speech detection \citewpar{gitari2015lexicon}. Over the last 20 years, social media networks have become a rich data source for sentiment analysis \citewpar{clarke-grieve-2017-dimensions,tian-etal-2017-facebook}. Extensive research has been done for sentiment analysis of monolingual corpora such as English \citewpar{10.1145/1014052.1014073,Wiebe2005,jiang-etal-2019-challenge}, Russian \citewpar{rogers-etal-2018-rusentiment}, German \citewpar{cieliebak-etal-2017-twitter}, Norwegian \citewpar{maehlum-etal-2019-annotating} and Indian languages \citewpar{agrawal-etal-2018-beating,priya-etal-2020-senti-comparative}. In initial research workds, n-gram features were used widely for classification of sentiments \citewpar{kouloumpis2011twitter}. However recently, due to readily available data on social media, these traditional techniques have been replaced by deep neural network techniques. \citewopar{patwa2020sentimix} conducted sentiment analysis on code-mixed social media text for Hindi-English and Spanish-English languages. However, sentiment analysis in Dravidian languages is under-studied. 

The anonymous and consequence-free nature of social media posts has proliferated the use of aggressive, hateful or offensive language online. This downturn has encouraged the development of automatic moderation systems. These systems if trained on proper data can help detect aggressive speech thus moderating spiteful content on a public platform. Collection of such data has become a crucial part of social media analysis. To facilitate the researchers working on these problems, there have been shared tasks conducted on aggression identification in social media \citewpar{kumar-etal-2018-benchmarking} and offensive language identification \citewpar{zampieri-etal-2019-predicting} by providing necessary datasets. As English is a commonly used language on social media, a significant amount of research goes into the identification of offensive English text. However, many internet users prefer the use of their native languages. This has given rise to the development of offensive language identification dataset in Arabic, Danish, Greek, and Turkish languages \citewpar{zampieri-etal-2020-semeval}. Inspired by this we developed resources for offensive language identification for Dravidian languages. 

In the past few years, cheaper internet and increased use of smartphones have significantly increased social media interaction in code-mixed native languages. Dravidian language speakers (who are often bilingual with English as it is an official language in India) with a population base of 215 million \footnote{\url{https://www.britannica.com/topic/Dravidian-languages}} contribute to large portion of such interactions. Hence, there is an ever-increasing need for the analysis of code-mixed text in Dravidian languages. However, the number of freely available code-mixed dataset \citewpar{chakravarthi2016,chakravarthi-code-mix-survey} are still limited in number, size, and  availability. Towards building language identification (LID) systems in code-mixed languages, \citewopar{8447784} developed a Kannada-English dataset containing English and Kannada text with word-level code-mixing. Also, they employed a stance detection system to detect stance in Kannada-English code-mixed text (on social media) using sentence embeddings. \citewopar{8554835} have used distributed representations for sentiment analysis of Kannada-English code-mixed texts through neural networks, which had three tags: Positive, Negative and Neutral. However, the dataset for Kannada was not readily available for research purposes. To give motivation for further research we conducted \citewpar{chakravarthi2020overview,dravidiansentiment-acm,10.1145/3441501.3441517,chakravarthi-etal-2021-findings-shared-task} a shared task that provided Tamil-English, Kannada-English, and Malayalam-English code-mixed datasets using which participants trained models that identify the sentiments (task A) and offensive classes (task B) in both the languages. 

Most of the recent studies on sentiment analysis and offensive language identification have been conducted on high-resourced languages from social media platforms. Models trained on such highly resourced monolingual data have succeeded in predicting sentiment and offensiveness. However, with the increased social media usage of bilingual users, a system trained on under-resourced code-mixed data is needed. In spite of this need, no large datasets for Tamil-English, Kannada-English and Malayalam-English are  available. Hence, inspired by \citewopar{severyn-etal-2014-opinion}, we collected and created a code-mixed dataset from YouTube. In this work, we describe the process of corpora creation for under-resourced Dravidian languages from YouTube comments. This is an extension of two workshop papers \citewpar{chakravarthi-etal-2020-senti-malayalam,chakravarthi-etal-2020-senti-tamil} and shared tasks \citewpar{dravidiansentiment-acm}. We present DravidianCodeMix corpora for Tamil-English (40,000+ comments), Kannada-English (7,000+ comments) and Malayalam-English (nearly 20,000 comments)  with manually annotated labels for sentiment analysis and offensive language identification. We used Krippendorff's alpha to calculate agreement amongst annotators. We made sure that each comment is annotated by at least three annotators and made the labelled corpora freely available for research purpose. For bench marking, we provided baseline experiments and results on 'DravidianCodeMix' corpora using machine learning models. 

\begin{figure}
  \includegraphics[trim=2cm 5cm 1cm 5cm, width=\textwidth]{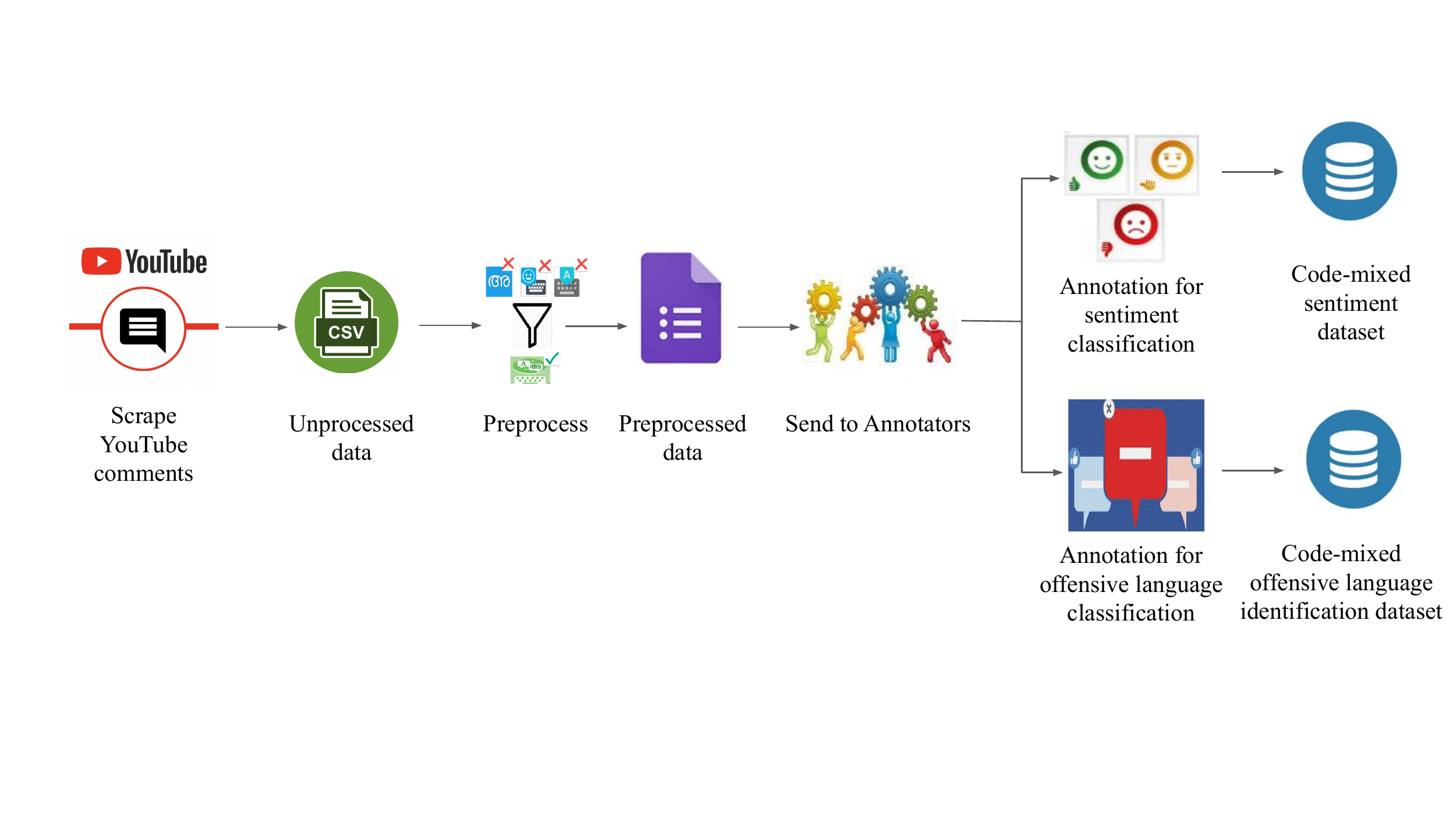}
  \caption{Data collection process.}
  \label{fig:flow}
\end{figure}

\begin{figure*}
  \includegraphics[width=\textwidth]{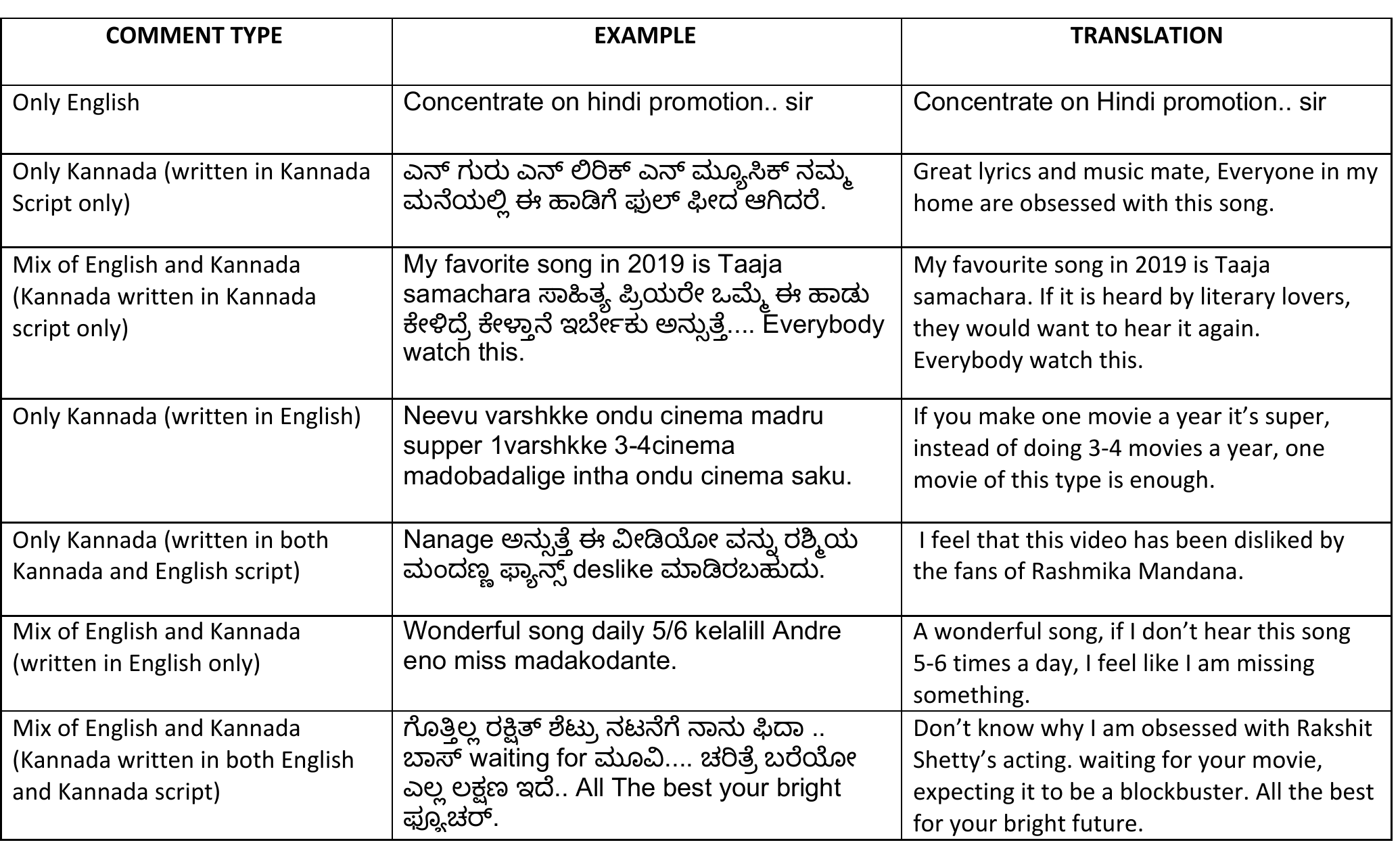}
  \caption{Examples of code mixing in Kannada dataset.}
  \label{fig:example-codemix-kn}
\end{figure*}

\begin{figure*}
  \includegraphics[width=\textwidth]{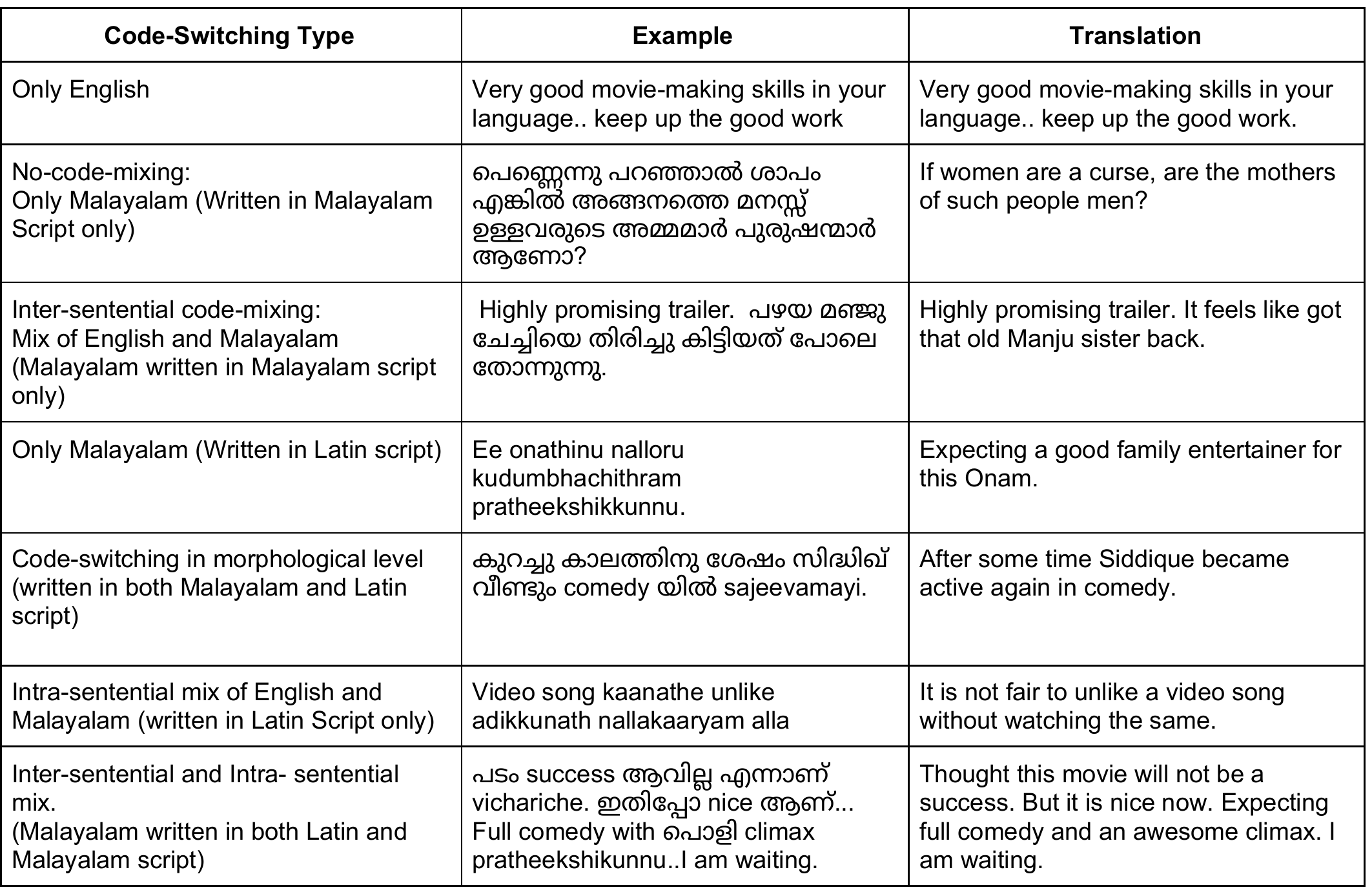}
  \caption{Examples of code mixing in Malayalam dataset.}
  \label{fig:example-codemix-ml}
\end{figure*}
\section{Raw Data} 
Online media, for example, Twitter, Facebook or YouTube, contain quickly changing data produced by millions of users that can drastically alter the reputation of an individual or an association. This raises the significance of programmed extraction of sentiments and offensive language used in online social media. YouTube is one of social media which is getting popular in the Indian subcontinent because of the wide range of content available from the platform such as songs, tutorials, product reviews, trailers and so on. YouTube allows users to create content and other users to comment on the content. It allows for more user-generated content in under-resourced languages. Hence, we chose YouTube to extract comments to create our dataset.  We chose movie trailers as the topic to collect data because movies are quite popular among the Tamil, Malayalam, and Kannada speaking populace. This increases the chance of getting varied views on one topic. Figure \ref{fig:flow} shows the overview of the steps involved in creating our dataset.

We compiled the comments from different film trailers of Tamil, Kannada, and Malayalam languages from YouTube in the year 2019. The comments were gathered using \textit{YouTube Comment Scraper tool}\footnote{\url{https://github.com/philbot9/youtube-remark scraper}}. We utilized these comments to make the datasets for sentiment analysis and offensive language identification with manual annotations. We intended to collect comments that contain code-mixing at various levels of the text, with enough representation for each sentiment and offensive language classes in all three languages. It was a challenging task to extract the necessary text that suited our intent from the comment section, which was further complicated by the presence of remarks in other non-target languages.  As a part of the preprocessing steps to clean the data, we utilized \textit{langdetect library} \footnote{\url{https://pypi.org/venture/langdetect/}} to tell different languages apart and eliminate the unintended languages. Examples of code-mixing in Tamil, Kannada and Malayalam corpora are shown in Figure \ref{fig:example-codemix-ta}, Figure \ref{fig:example-codemix-kn}, and Figure \ref{fig:example-codemix-ml} along with their translations in English. By keeping data privacy in mind, we made sure that all the user-related information is removed from the corpora. As a part of the text-preprocessing, we removed redundant information such as URL.

Since we collected corpora from social media, our corpora contain different types of real-world code-mixed data. Inter-sentential switching is characterised by change of language between sentences where each sentence is written or spoken in one language. Intra-sentential switching occurs within a single sentence, say one of the clause is in one language and the other clause is in the second language. Our corpora contains all forms of code-mixing ranging from purely monolingual texts in native languages to mixing of scripts, words, morphology, inter-sentential and intra-sentential switches. We retained all the instances of code-mixing to faithfully preserve the real-world usage.
\begin{figure*}
  \includegraphics[width=\textwidth]{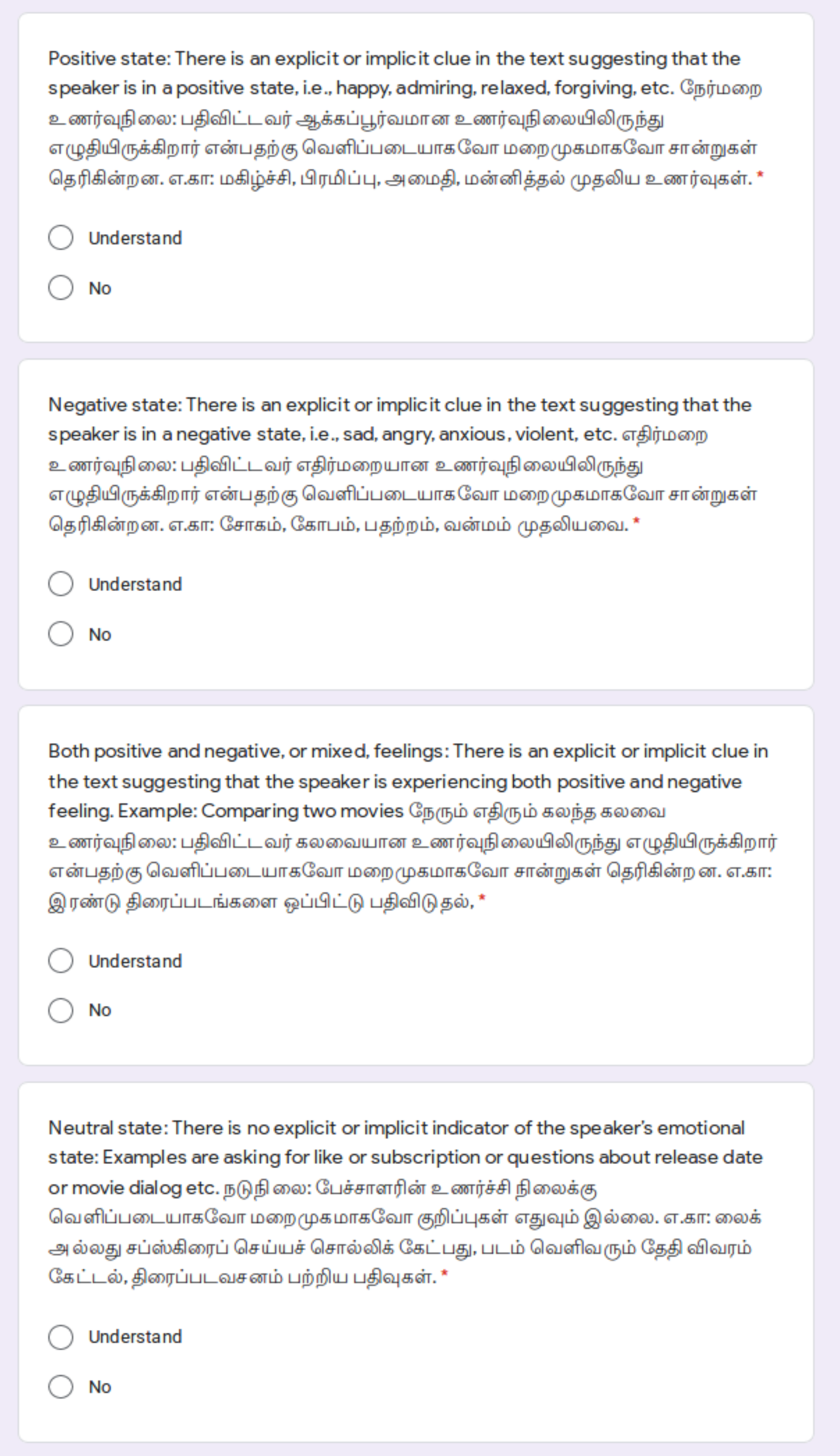}
  \caption{Example Google Form with annotation instructions for sentiment analysis}
  \label{fig:firstpageannotation}
\end{figure*}
\section{Methodology of Annotation}
We create our corpora for two tasks, namely, sentiment analysis and offensive language identification. We anonymized the data gathered from Youtube in order to protect user privacy.
\subsection{Annotation Process}
In order to find volunteers for the annotation process, we contacted students in Indian Institute of Information Technology and Management-Kerala for Malayalam, Indian Institute of Information Technology-Tiruchirapalli and Madurai Kamaraj University for Tamil. For Kannada, we contacted students in Visvesvaraya College of Engineering, Bangalore University. The student volunteer annotators received the link to a Google Form and did the annotations on their personal computers. The authors’ family members also volunteered to annotate the data. We created Google Forms to gather annotations from annotators. Information on gender, education background and medium of schooling were collected to know the diversity of the annotators. The annotators were cautioned that the user remarks may have hostile language. They were given a provision to discontinue with the annotation process in case the content is too upsetting to deal with. They were asked not to be partial to a specific individual, circumstance or occasion during the annotation process. Each Google form had been set to contain up to 100 comments and each page was limited to contain 10 comments. The annotators were instructed to agree that they understood the scheme before they were allowed to proceed further. The annotation setup involved three stages. To begin with, each sentence was annotated by two individuals. In the second step, the data was included in the collection if both the annotations agreed. In the event of contention, a third individual was asked to annotate the sentence. In the third step, in the uncommon case that all the three of them disagreed, at that point, two additional annotators were brought in to label the sentences. Each form was annotated by at least three annotators.
\subsection{Sentiment Analysis}
\begin{figure*}
  \includegraphics[width=\textwidth]{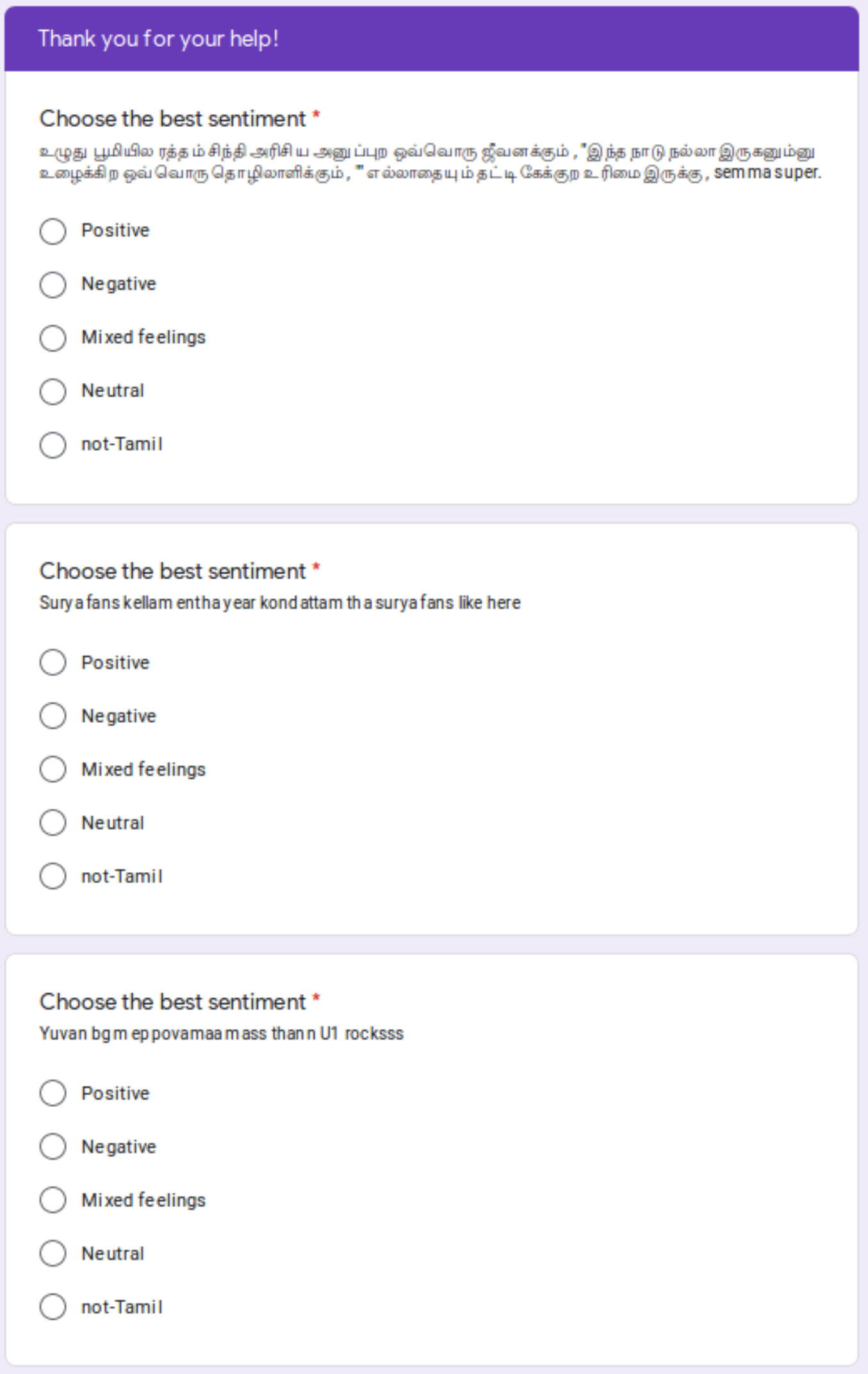}
  \caption{Examples from the first page of the Google form for sentiment analysis}
  \label{fig:senti-example}
\end{figure*}

For sentiment analysis, we followed the methodology taken by \citewopar{chakravarthi-etal-2020-senti-tamil}, and involved at least three annotators to label each sentence. The following annotation schema was given to the annotators in English and Dravidian languages.
\begin{itemize}
    \item \textbf{Positive state:} Comment contains an explicit or implicit clue in the content recommending that the speaker is in a positive state. 
    \item \textbf{Negative state:} Comment contains an explicit or implicit clue in the content recommending that the speaker is in a negative state. 
    \item \textbf{Mixed feelings:} Comment contains an explicit or implicit clue in both positive and negative feeling.
    \item \textbf{Neutral state:} Comment does not contain an explicit or implicit indicator of the speaker’s emotional state.
    \item \textbf{Not in intended language:} If the comment is not in the intended language. For example, for Tamil, if the sentence does not contain Tamil written in Tamil script or Latin script, then it is not Tamil.
\end{itemize}
Figures \ref{fig:firstpageannotation} and \ref{fig:senti-example} show the sample Google Forms for general instructions and sentiment analysis respectively.
\subsection{Offensive Language Identification}
We constructed offensive language identification dataset for Dravidian languages at different levels of complexity following the work of \citewopar{zampieri-etal-2019-predicting}. More generally we expand this expand to a three-level hierarchical annotation schema. We added a new category \textbf{Not in intended language} to account for comments written in a language other than the intended language. Examples for this are the comments written in other Dravidian languages using Roman script. To simplify the annotation decisions, we split offensive language categories into six labels. 
\begin{itemize}
    \item  \textbf{Not Offensive}: Comment does not contain offence or profanity.
    \item  \textbf{Offensive Untargeted}: Comment contains offence or profanity not directed towards any target. These are the comments which contain unacceptable language without targeting anyone. 
    \item  \textbf{Offensive Targeted Individual}: Comment contains offence or profanity which targets an individual.
    \item  \textbf{Offensive Targeted Group}: Comment contains offence or profanity which targets a group or a community.
    \item  \textbf{Offensive Targeted Other}: Comment contains offence or profanity which does not belong to any of the previous two categories (e.g. a situation, an issue, an organization or an event). 
    \item  \textbf{Not in indented language}:  If the comment is not in the intended language. For example, in Tamil task, if the sentence does not contain Tamil written in Tamil script or Latin script, then it is not Tamil.
\end{itemize}
\begin{figure*}
  \includegraphics[width=\textwidth]{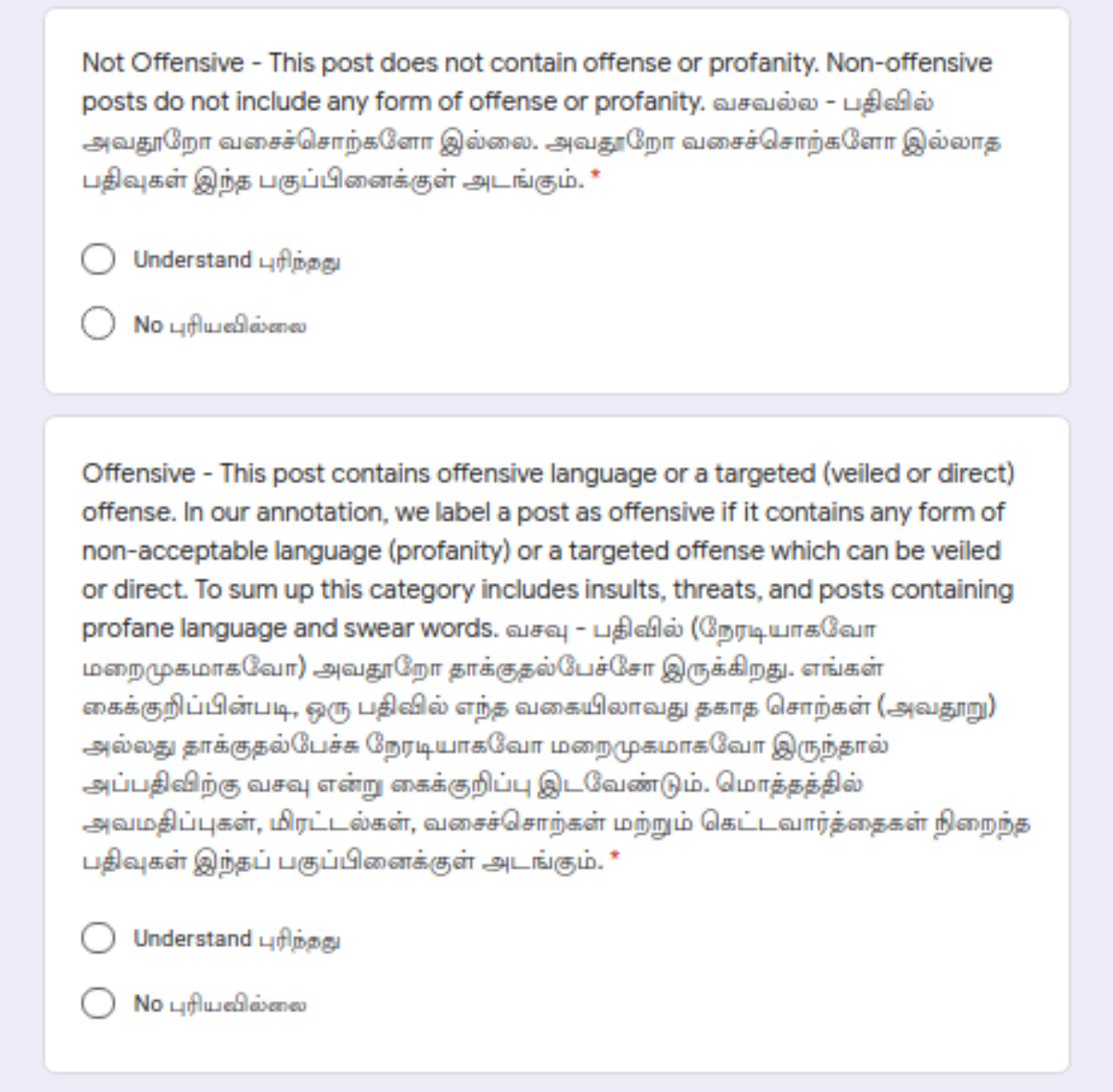}
  \caption{Example Google Form with annotation instructions for offensive language identification}
  \label{fig:offen1}
\end{figure*}
\begin{figure*}
\centering
  \includegraphics[width=\textwidth]{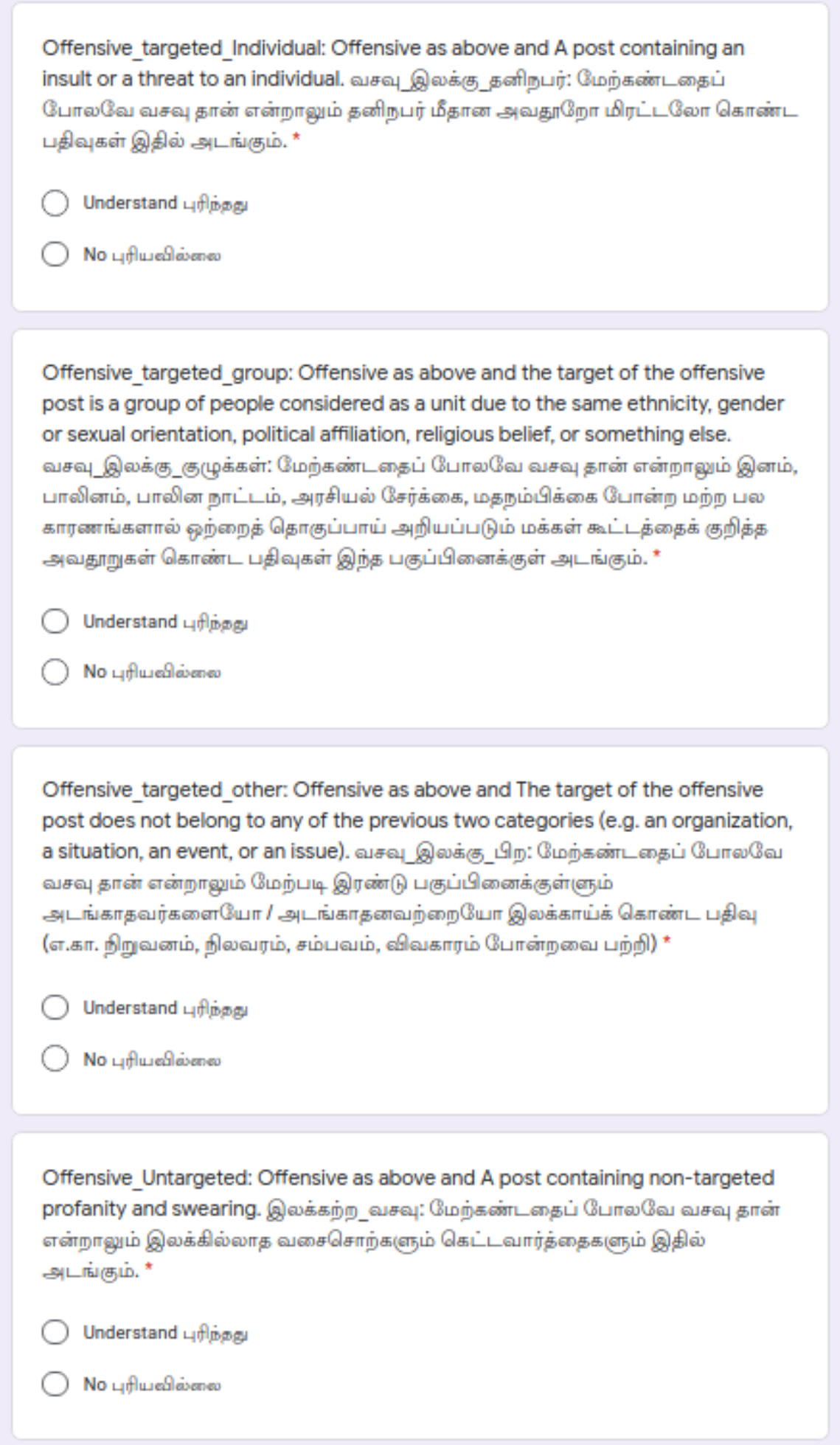}
  \caption{Example Google Form with annotation instructions for offensive language identification}
  \label{fig:offen2}
\end{figure*}
\begin{figure*}
  \includegraphics[width=\textwidth]{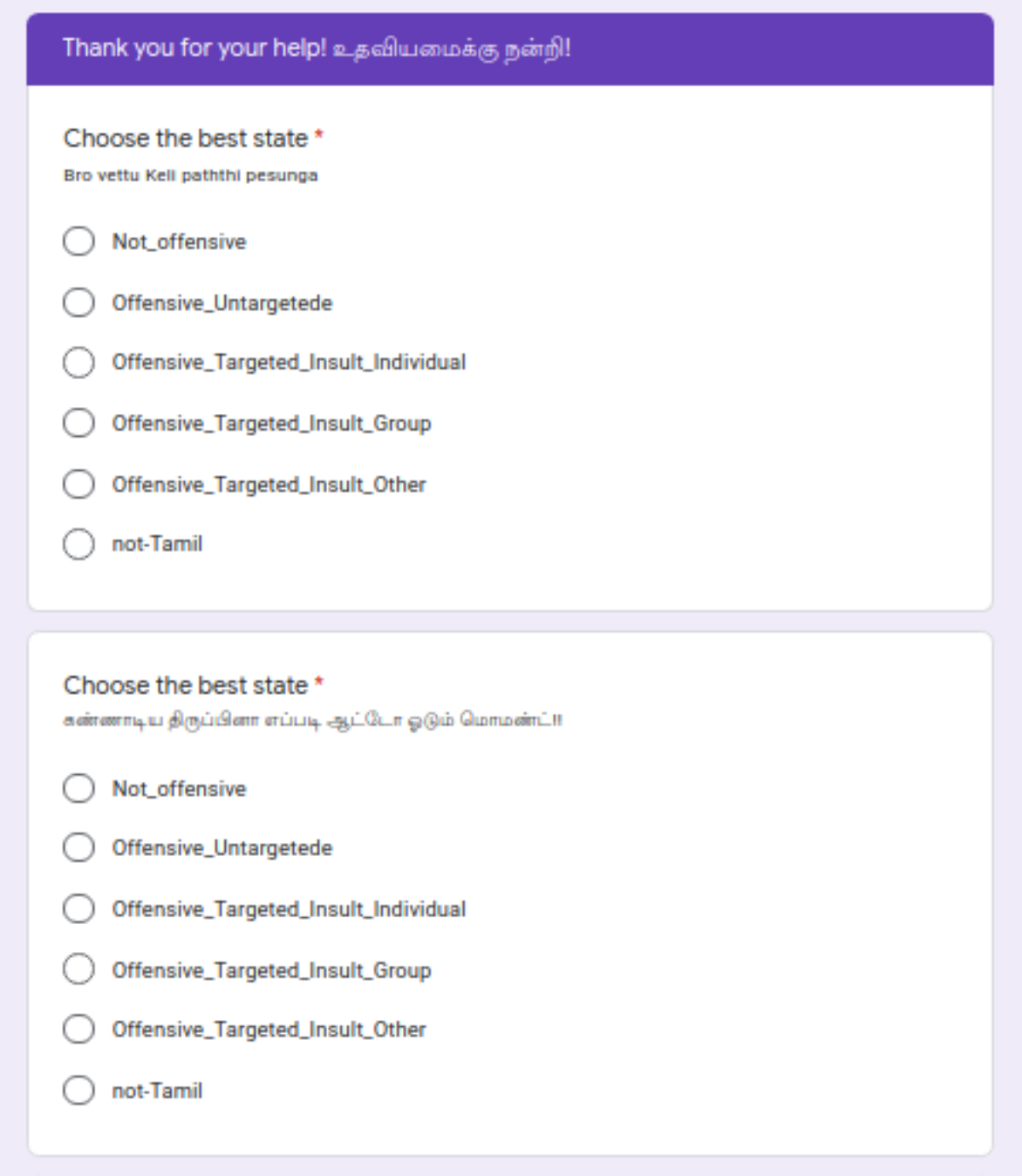}
  \caption{Examples from the first page of the Google Form for offensive language identification}
  \label{fig:offen3}
\end{figure*}
Examples of the Google Forms in  English and native language for offensive language identification task are given in Figure \ref{fig:offen1}, Figure \ref{fig:offen2}, and Figure \ref{fig:offen3}.
\begin{table}[!htb]  
\begin{center} 
\renewcommand{\tabcolsep}{1.5mm}
\normalsize
\begin{tabular}{|l|l|c|c|c|}
\hline
Language & & Tamil & Malayalam & Kannada\\
\hline
Gender & Male &9 & 2  &2  \\
& Female &2  & 4  & 3\\
&Non-binary &0&0&0 \\
\hline
Higher Education & Undegraduate & 2 & 0 & 1 \\
& Graduate & 2 & 0 & 2 \\
& Postgraduate & 7 & 6 & 2 \\
\hline
Medium of Schooling & English & 6 & 5  & 4 \\
& Native language & 5 & 1 &  1\\
\hline
Total &  &11 & 6 &5\\
\hline
\end{tabular}
\caption{Annotators Statistics for Sentiment Analysis} 
\label{tab:annotators} 
\end{center} 
\end{table}
\begin{table}[!htb]  
\begin{center} 
\renewcommand{\tabcolsep}{1.5mm}
\normalsize
\begin{tabular}{|l|l|c|c|c|}
\hline
Language & & Tamil & Malayalam & Kannada\\
\hline
Gender & Male &6 & 2  &3  \\
& Female &6  & 4  & 2\\
&Non-binary &0&0&0 \\
\hline
Higher Education & Undegraduate & 2 & 0 & 0 \\
& Graduate & 5 & 0 & 3 \\
& Postgraduate & 5 & 6 & 2 \\
\hline
Medium of Schooling & English & 6 & 5  & 3 \\
& Native language & 7 & 1 &  2\\
\hline
Total &  &12 & 6 &5\\
\hline
\end{tabular}
\caption{Annotators Statistics for Offensive Language Identification} 
\label{tab:annotators2} 
\end{center} 
\end{table}
Once the Google Form was ready, we sent it out to an equal number of males and females to enquire their willingness to annotate. We got varied responses from them and so our distribution of male and female annotators involved in the task are different. From Table \ref{tab:annotators}, we can see that only two female annotators volunteered to contribute for Tamil while there were more female annotators for Malayalam and Kannada. For offensive language identification, we can see that there is a balance in gender from Table \ref{tab:annotators2}. The majority of the annotators have received postgraduate level of education. We were not able to find volunteers of non-binary gender to annotate our dataset. All the annotators who volunteered to annotate the Tamil-English, Kannada-English and Malayalam-English datasets had bilingual proficiency in the respective code-mixed pairs and they were prepared to take up the task seriously.  From Table \ref{tab:annotators} and \ref{tab:annotators2}, we can observe that the majority of the annotators' medium of schooling is English even though their mother tongue is Tamil, Kannada or Malayalam. For Kannada and Malayalam languages only one annotator from each language received their education through the medium of their native language. Although the medium of education of the participants was skewed towards the English language, we were careful it would not affect the annotation task by ensuring that all of them are fully proficient in using their native language. 

A sample form (first assignment) was annotated by experts and a gold standard was created. We manually compared the gold standard annotations with the volunteer submission form. To control the quality of annotation, we eliminated the annotators whose label assignments in the first form were not good. For instance, if the annotators showed an unreasonable delay in responding or if they labelled all sentences with the same label or if more than fifty annotations in a form were wrong, we eliminated those contributions.  A total of 22 volunteers and 23 volunteers, for sentiment analysis and offensive language identification tasks respectively, were involved in the process. Once they filled up the Google Form, 100 sentences were sent to them. If an annotator offered to volunteer more, the next Google Form was sent to them with another set of 100 sentences and in this way each volunteer chose to annotate as many sentences from the corpus as they wanted. 
\begin{table}[!htb]  
\begin{center} 
\renewcommand{\tabcolsep}{1.5mm}
\normalsize
\begin{tabular}{|ccccc|}
\hline
& \multicolumn{2}{c}{Sentiment Analysis} & \multicolumn{2}{c}{Offensive Language Identification }\\
\hline
& Nominal & Ordinal & Nominal & Ordinal \\
\hline
Tamil & 0.6735 & 0.6534 & 0.7452 & 0.7634 \\
Malayalam & 0.8753 & 0.8463 & 0.8345 & 0.8374 \\
Kannada & 0.7356 & 0.7465 & 0.8456 & 0.8443 \\
\hline
\end{tabular} 
\caption{Inter-annotator agreement in Krippendorff's alpha} 
\label{tab:annotators3} 
\end{center} 
\end{table}
\subsection{Inter-annotator agreement}
Inter-annotator agreement is a measure of the extent to which the annotators agree in their rating. This is necessary to ensure that the annotation scheme is consistent and that different raters are able to assign the same sentiment label to a given comment. There are two questions related to inter-annotator agreement: How do the annotators agree or disagree in their annotation? How much of the observed agreement or disagreement among the annotators might be due to chance? While the percentage of agreement is fairly straightforward, answering the second question involves defining and modelling what chance is and how to measure the agreement due to chance. There are different inter-annotator agreement measures that are intended to answer this in order to measure the reliability of the annotation. We utilized \textbf{Krippendorff's alpha $(\alpha)$} \citewpar{krippendorff5} to gauge the agreement between annotators because of the nature of our annotation setup. Krippendorff's alpha is a rigorous statistical measure that accounts for incomplete data and, consequently, does not require every annotator to annotate every sentence. It is also a measure that considers the level of disagreement between the anticipated classes, which is critical in our annotation scheme. For example, if the annotators differ among \textbf{Positive} and \textbf{Negative} class, this difference is more genuine than when they differ between \textbf{Mixed feelings} and \textbf{Neutral state}. $\alpha$ is sensitive to such disagreements. $\alpha$ is characterized by:
\begin{equation}
    \alpha = 1 - \frac{D_o}{D_e}
\end{equation}
$D_o$ is the observed disagreement between sentiment labels assigned by the annotators and $D_e$ is the disagreement expected when the coding of sentiments can be attributed to chance rather than due to the inherent property of the sentiment itself. 
\begin{equation}
    D_o = \frac{1}{n}\sum_{c}\sum_{k}o_{ck\;metric}\;\delta^2_{ck}
\end{equation}
\begin{equation}
D_e = \frac{1}{n(n-1)} \sum_{c}\sum_{k}n_c \; .\;n_{k\;metric}\,\delta^2_{ck}
\end{equation}
Here $o_{ck}\;n_c\;n_k\;$ and $n$ refer to the frequencies of values in the coincidence matrices and $metric$ refers to any metric or level of measurement such as nominal, ordinal, interval, ratio and others. Krippendorff's alpha applies to all these metrics. We used nominal and ordinal metric to calculate inter-annotator agreement. The range of $\alpha$ is between `0' and `1', $1 \ge \alpha \ge 0$. When $\alpha$ is `1' there is perfect agreement between the annotators and when `0' the agreement is entirely due to chance. Care should be taken in interpreting the reliability of the results shown by Krippendorf's alpha because reliability basically measures the amount of noise in the data. However, the location of noise and the strength of the relationship measured will interfere with the reliability of the estimate. It is customary to require $\alpha$  $\geq$ .800. A reasonable rule of thumb that allows for tentative conclusions to be drawn requires $0.67 \leq \alpha \leq 0.8 $ while $\alpha \ge$ .653 is the lowest conceivable limit. We used \textit{nltk}\footnote{\url{https://www.nltk.org/}} for calculating Krippendorff's alpha $(\alpha)$. The results of inter-annotator agreement between our annotators for different languages on both sentiment analysis and offensive language identification tasks are shown in Table \ref{tab:annotators3}. 

\section{Corpus Statistics}
Table \ref{tab:corp_stat1} and Table \ref{tab:corp_stat2} show the text statistics (number of words, vocabulary size, number of comments, number of sentences, and average number of words per sentences) for sentiment analysis and offensive language identification for Tamil, Malayalam and Kannada. The Tamil dataset had the highest number of samples while Kannada had the least on both the tasks. On average, each comment contained only one sentence.

\begin{figure}
  \includegraphics[width=\textwidth]{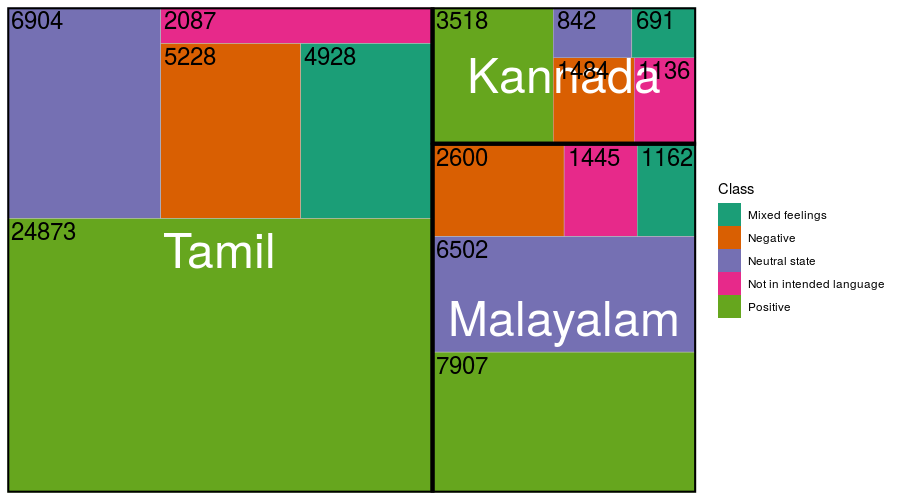}
  \caption{Treemap for comparing sentiment classes across Tamil, Malayalam and Kannada}
  \label{fig:class_dist_sent}
\end{figure}

\begin{figure*}
  \includegraphics[width=\textwidth]{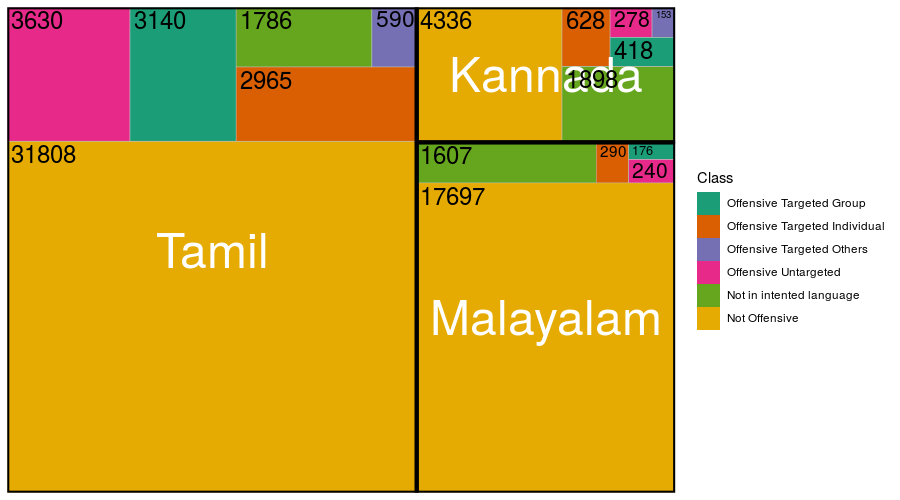}
  \caption{Treemap for comparing offensive classes across Tamil, Malayalam and Kannada}
  \label{fig:class_dist_off}
\end{figure*}

\begin{figure*}
  \includegraphics[width=\textwidth]{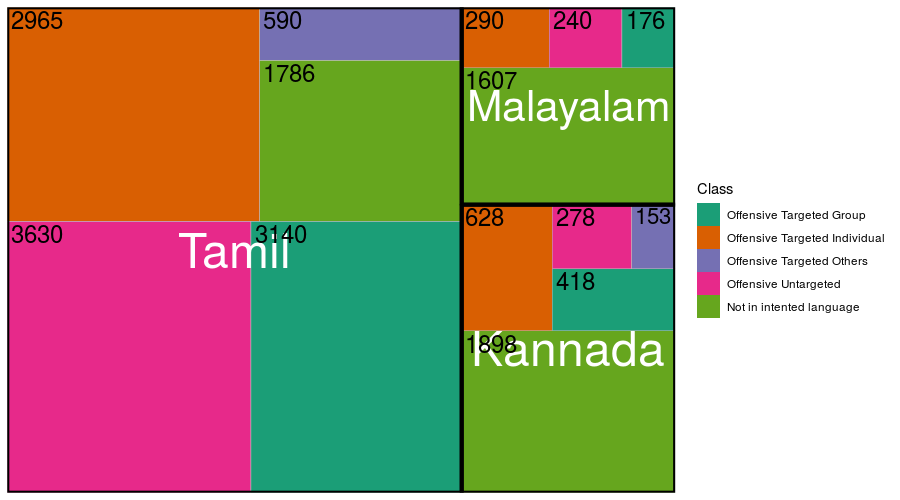}
  \caption{Treemap for comparing offensive classes (excluding Not Offensive class) across Tamil, Malayalam and Kannada}
  \label{fig:class_dist_just_off}
\end{figure*}

Table \ref{tab:senti-data_distribution} and Table \ref{tab:off-data_distribution} show the class distribution across Tamil, Malayalam and Kannada in sentiment analysis and offensive language identification tasks. Furthermore, tree-maps in Figure \ref{fig:class_dist_sent} and Figure \ref{fig:class_dist_off} depict the comparative analysis of distribution of sentiment and offensive classes across languages. Figure \ref{fig:class_dist_sent} illustrates that there are more number of samples labelled "Positive" than any other class in all the languages. While the disparity between "Positive" and other classes is large in Tamil, it is not the case with Malayalam and Kannada. In Malayalam, "Neutral state" is the second-largest class in terms of distribution; 6,502 number of comments labelled "Neutral state" could mean that most of the comments in Malayalam are vague remarks as the sentiment behind them is unknown. On the other hand, Kannada has the least number of "Neutral state" class. 
Figure \ref{fig:class_dist_off} shows that all languages have not-offensive class in the majority. In the case of Tamil, 71\% of the total comments are not offensive, while Malayalam has 85\% non-offensive comments. But there is no consistent trend observable amongst offensive classes across the languages shown in Figure  \ref{fig:class_dist_just_off}. In the case of Tamil, 60\% of the offensive comments are targeted (group or individual). Similar trends are seen in the case of Malayalam (66\%) and Kannada (79\%). Absence (Malayalam) or least (Tamil, Kannada) number of targeted other category comments points to the fact that most of the offensive comments are targeted towards either an individual or a group. In the case of Kannada, it is interesting to see that 24\% out of total comments are in a language other than Kannada. This could mean that a Kannada movie gets a significant amount of audience who are not native Kannada speakers or that Kannada speakers tend to use more languages other than English to generate code-mixed content online. 
\begin{table*}[!htb]  
\begin{center} 
\renewcommand{\tabcolsep}{1.5mm}

\normalsize
\begin{tabular}{|l|r|r|r|}
\hline
Language & Tamil  &Malayalam &Kannada\\
\hline
Number of words & 513,311 & 224,207 & 65,002\\
Vocabulary Size & 94,928 & 57,566 & 20,665\\
Number of comments & 44,020 & 19,616 & 7,671\\
Number of sentences & 52,750 & 24,014 & 8,472\\
Average number of words per sentence & 11 & 11 & 8\\
Average number of sentences per comment & 1 & 1 & 1\\
\hline
\end{tabular} 
\caption{Corpus statistics for Sentiment Analysis} 
\label{tab:corp_stat1} 
\end{center} 
\end{table*}

\begin{table*}[!htb]  
\begin{center} 
\renewcommand{\tabcolsep}{1.5mm}
\normalsize
\begin{tabular}{|l|r|r|r|}
\hline
Language & Tamil  &Malayalam &Kannada\\
\hline
Number of words & 511,734 & 202,134 & 65,702\\
Vocabulary size & 94,772 & 40,729 & 20,796\\
Number of comments & 43,919 & 20,010 & 7,772\\
Number of sentences & 52,617 & 23,652 & 8,586\\
Average number of words per sentence & 11 & 10 & 8\\
Average number of sentences per comment & 1 & 1 & 1\\
\hline
\end{tabular} 
\caption{Corpus statistics for Offensive Language Identification} 
\label{tab:corp_stat2} 
\end{center} 
\end{table*}


\begin{table}[!htb]  
\begin{center} 
\renewcommand{\tabcolsep}{1.5mm}
\normalsize
\begin{tabular}{|l|r|r|r|}
\hline
Class & Tamil & Malayalam & Kannada \\
\hline
Negative & 5,228 (11.87 \%) & 2,600 (13.25 \%) & 1,484 (19.34 \%)\\
Not in intended language & 2,087 (4.74 \%) & 1,445 (736 \%) & 1,136 (14.80 \%)\\
Neutral state & 6,904 (15.68 \%) & 6,502 (33.14 \%) & 842 (10.97 \%)\\
Mixed feelings & 4,928 (1119 \%) & 1,162 (5.92 \%) & 691 (9.00 \%)\\
Positive & 24,873 (56.50 \%) & 7,907 (40.30 \%) & 3,518 (45.86 \%)\\
\hline
Total & 44,020 & 19,616 & 7,671\\
\hline
\end{tabular} 
\caption{Sentiment Analysis Dataset Distribution} 
\label{tab:senti-data_distribution} 
\end{center} 
\end{table}
\begin{table}[!htb]  
\begin{center} 
\renewcommand{\tabcolsep}{1.5mm}
\normalsize
\begin{tabular}{|l|r|r|r|}
\hline
Class &Tamil &Malayalam&  Kannada \\
\hline
Not Offensive & 31,808 (72.42 \%) & 17,697 (88.44 \%) & 4,336 (55.79 \%) \\
O-Untargeted & 3,630 (8.26 \%) & 240 (1.19 \%) & 278 (3.57 \%)\\
O-Targeted Individual & 2,965 (6.75 \%) & 290 (1.44 \%) & 628 (8.08 \%)\\
O-Targeted Group & 3,140 (7.14 \%) & 176 (0.87 \%) & 418 (5.37 \%)\\
O-Targeted Others & 590 (1.34 \%) & - & 153 (1.96 \%) \\
Not in indented lang & 1,786 (4.06 \%) & 1,607 (8.03 \%) & 1,898 (24.42 \%)\\
\hline
Total & 43,919 & 20,010 & 7,772\\
\hline
\end{tabular} 
\caption{Offensive language Identification Dataset Distribution. O-Offensive. O-Untargeted: Offensive Untargeted. } 
\label{tab:off-data_distribution} 
\end{center} 
\end{table}
Our datasets are stored in tab separated files. The first column of the tsv file contains the comments from YouTube and the second column has the final annotation. 
\section{Difficult Examples}
The social media comments that form our dataset are code-mixed showing a mixture of Dravidian languages and English. This poses a few major difficulties while annotating the sentiments and offensive language categories on our dataset. Dravidian languages are under-resourced languages and the mixing of scripts makes the annotation task difficult since the annotators must have learned both the scripts, be familiar with how English words are modified to native phonology and how the meaning of certain English words have a different meaning in the given local language. Reading and understanding the code mixed text often with non-standardised spelling is difficult. Moreover, we have created the annotation labels with the help of volunteer annotators for three languages (not just one language). It is challenging and time consuming to collect this much amount of data from bilingual, volunteer annotators from three different language groups.

While annotating, it was found that some of the comments were ambiguous in conveying the right sentiment of the viewers. Hence the task of annotation for sentiment analysis and offensive language identification seemed difficult. The problems include the comparison of the movie with movies of same or other industries, expression of opinion of different aspects of the movie in the same sentence. Below are a few examples of such comments and details of how we resolved those issues are provided. In this section, we talk about some examples from Tamil language that were difficult to annotate. 
\begin{itemize}
    \item \textbf{\color{blue}Enakku iru mugan {\color{red}trailer} gnabagam than varuthu} \-- \textit{All it reminds me of is the trailer of the movie Irumugan}. Not sure whether the speaker enjoyed Irumugan trailer or disliked it or simply observed the similarities between the two trailers. The annotators found it difficult to identify the sentiment behind the comment consistently. 
    \item \textbf{\color{blue}Rajini ah vida akshay {\color{red} mass} ah irukane} \-- \textit{Akshay looks more amazing than Rajini}. Difficult to decide if it is a disappointment that the villain looks better than the hero or a positive appreciation for the villain actor. Some annotators interpreted negative sentiment while some others took it as positive.
    \item \textbf{\color{blue} Ada dei nama sambatha da dei
} \-- \textit{I wonder, Is this our sampath? Hey!.} Conflict between neutral and positive. 
    \item \textbf{\color{blue}Lokesh kanagaraj {\color{red}movie} naalae.... {\color{red}English Rap....Song} vandurum} \-- \textit{If it is a movie of Lokesh kanagaraj, it always has an English rap song}. Ambiguous sentiment.
    \item \textbf{\color{blue}Ayayo bigil aprm {\color{red}release} panratha {\color{red}idea} iruka lokesh {\color{red}gaaru}} \-- \textit{Oh Dear! Are you even considering releasing the movie Bigil, Mr.Lokesh?}. This comment has a sinlge word `garu'\footnote{Telugu word for Mr} which is a non-Tamil , non-English word borrowed from Telugu language which is a politeness marker. However, in this context the speaker uses the word sarcastically to insult the director because of the undue delay in releasing the movie. The annotators were inconsistent in interpreting this as offensive or not-Tamil.
    \item \textbf{\color{blue}{\color{red}No of dislikes} la theriyudhu, idha yaru {\color{red}dislike} panni irrupanga nu} \-- \textit{It is obvious from the number of dislikes as to who would have disliked this (trailer).} The comment below the trailer of a movie which talks about the caste issues in contemporary Tamil society. Based on the content of the trailer, the speaker offensively implies that the scheduled caste people are the ones who would have disliked the movie and not other people. Recognising the offensive undercurrent in a seemingly normal comment is difficult and hence these examples complicate the annotation process.
\end{itemize}
According to the instructions, questions about music director, movie release date and comments containing speaker's remarks about the date and time of watching the video should be treated as belonging to neutral class. However the above examples show that some comments about the actors and movies can be ambiguously interpreted as neutral or positive or negative. We found annotator disagreements in such sentences. Below, we give similar examples from Malayalam.
\begin{itemize}
\item \textbf{\color{blue}\textit{{\color{red}Realistic} bhoothanghalil ninnu oru vimochanam pratheekshikkunnu}} \--\textit{Hoping for a deliverance from realistic demons.} No category of audience can be pleased simultaneously. The widespread opinion is that the Malayalam film industry is advancing with more realistic movies. Therefore a group of audience who is more fond of action or non-realistic movies are not satisfied with this culture of realistic movies. In this comment, the viewer is not insulting this growing culture, but expecting that the upcoming film is of his favourite genre. Hence we labelled it non-offensive.
\item \textbf{\color{blue}\textit{Ithilum valiya jhimikki kammal vannatha}} \-- \textit{There was an  even bigger `pendant earring'.} {\color{blue}\textit{ `Jhimikki kammal' }} was a trending song from a movie of the same actor mentioned here. The movie received huge publicity even before its release because of the song but it turned out to be a disappointment after its release. Thus the annotators got confused whether the comment is meant as an insult or not. But we concluded that the viewer is not offending the present trailer but marks his opinion as a warning for the audience to not judge the book by its cover.
\item \textbf{\color{blue}\textit{ Ithu kandittu nalla tholinja {\color{red}comedy}aayi thonniyathu enikku mathram aano?}}\-- \textit{Am I the only person here who felt this a stupid comedy?} The meaning of the Malayalam word mentioned here corresponding to the word ‘stupid’ varies with regions of Kerala. Hence the disparity in opinion between annotators who speaks different dialects of Malayalam was evident. Though in few regions it is offensive, generally it is considered as a byword for ‘bad’. 
\item \textbf{\color{blue}\textit{ aa {\color{red}cinema}yude peru kollam. Ithu Dileep ne udheshichanu,ayale mathram udheshichanu}} \--\textit{The name of that movie is good. It is named after Dileep and intended only for him.} It is quite obvious that there is a chance of imagining several different movie names based on the subjective predisposition of the annotator. As long as the movie name is unknown here, apparently no insult can be proved and there is no profane language used in the sentence either.
\item \textbf{\color{blue}\textit{Kanditt Amala Paul Aadai Tamil mattoru {\color{red}version} aanu ennu thonnunu}} \-- \textit{It looks like another version of Amala Paul's Tamil movie Aadai.} Here the viewer doubts the Malayalam movie `Helen' is similar to the Tamil movie `Aadai'. Though the movie `Aadai' was  positively received by viewers and critics, we cannot generalize and assume that this comment also as positive only because of this comparison. Hence we add it to the category of `mixed feeling'.
\item \textbf{\color{blue}\textit{Evideo oru {\color{red}Hollywood story} varunnilleee. Oru DBT.}} \--\textit{Somewhere there is a Hollywood storyline...one doubt.} This is also a comparison comment of that same movie `Helen' mentioned above. Nevertheless, here the difference is that the movie is compared with the Hollywood standard, which is well-known worldwide and is generally considered positive. Hence it is marked as a positive comment.
\item \textbf{\color{blue}\textit{{\color{red}Trailer} pole nalla {\color{red}story} undayal mathiyarinu.}}\--\textit{It was good enough to have a good story like the trailer.} Here viewer mentioned about two aspects of the movie viz: `trailer' and `story'. He appreciates the trailer but doubts the quality of the story at the same time. We considered this comment positive because it is clear that he enjoyed the trailer and conveys strong optimism for the movie.
\end{itemize}
\section{Benchmark Systems}
In this section, we report the results obtained in three languages for both the tasks in the corpora introduced above. Like many earlier studies, we  approach the tasks as text classification tasks. In order to provide a simple baseline, we applied several traditional machine learning algorithms such as Logistic Regression (LR), Support Vector Machine (SVM), Multinomial Naive Bayes (MNB), K-Nearest Neigbours (KNN), Decision Trees (DT) and Random Forests (RF) separately, for both sentiment analysis and offensive language detection on the code-mixed  datasets.

\begin{table}[!htb]  
\begin{center} 
\renewcommand{\tabcolsep}{1.5mm}
\normalsize
\begin{tabular}{|l|r|r|r|}
\hline
 &   Tamil & Malayalam & Kannada\\
\hline
Training & 35,220 & 15,694 & 6,136\\
Development & 4,398 & 1,960 & 767\\
Test & 4,402 & 1,962 & 768\\
\hline
Total & 44,020 & 19,616 & 7,671\\
\hline
\end{tabular} 
\caption{Train-Development-Test Data Distribution with 90\%-5\%-5\% train-dev-test split for Sentiment Analysis} 
\label{tab:train_data_distribution-senti} 
\end{center} 
\end{table}
\begin{table}[!htb]  
\begin{center} 
\renewcommand{\tabcolsep}{1.5mm}

\normalsize
\begin{tabular}{|l|r|r|r|}
\hline
 &   Tamil & Malayalam & Kannada\\
\hline
Training & 35,139 & 16,010 & 6,217\\
Development & 4,388 & 1,999 & 777\\
Test & 4,392 & 2,001 & 778\\
\hline
Total & 43,919 & 20,010 & 7,772\\
\hline
\end{tabular} 
\caption{Train-Development-Test Data Distribution with 90\%-5\%-5\% train-dev-test for Offensive Language Identification} 
\label{tab:train_data_distribution} 
\end{center} 
\end{table}
\subsection{Experiments Setup}
We used 90\%-5\%-5\% randomly sampled data split for training, development and test set for all the experimental setup. All the duplicated entries were removed from the dataset before the split to make test and development data truly unseen. All the experiments are tuned to the development set and tested on the test set.
\subsubsection{Logistic Regression (LR):}
LR is one of the base-line machine learning algorithms, which is also a probabilistic classifier used for the task of classification of data (\cite{article}). This is basically the transformed version of linear regression using the logistic function (\cite{park2013introduction}). Accordingly it takes the real-valued features as input which is later multiplied by a weight and the sum is fed to the sigmoid function $ \sigma  (z) $ also called the logistic function to obtain the class probability (\cite{shah2020comparative}). The decision is made based on the value set as threshold. Sigmoid function is as given below:
\begin{equation}
 \sigma  (z) =  \frac{\mathrm{1} }{\mathrm{1} + e^{-z} } 
\end{equation}
Logistic regression has a close relationship with neural networks as the latter can also be viewed as a stack of several LR classifiers (\cite{de-gispert-etal-2015-fast}). Unlike Naïve Bayes which is a generative classifier, LR is a discriminative classifier (\cite{ng2002discriminative}). While Naïve Bayes  holds strict conditional independence assumptions, LR is evidently more robust to correlated features (\cite{jin-pedersen-2018-duluth}). It means that when there are more than one features say F1,F2,F3 which are absolutely correlated, it will divide the weight W among the features as W1,W2,W3 respectively.

We evaluated the Logistic Regression model with L2 regularization to reduce overfitting. The input features are the Term Frequency Inverse Document Frequency (TF-IDF) values of up to 3 grams. This approach results in the model being trained only on this dataset without taking any pre-trained embeddings.
\subsubsection{Support Vector Machine (SVM):}
 Support Vector Machine are a powerful supervised machine learning algorithm used mainly for classification tasks and for regression as well. The goal of an SVM is to find the hyperplane in an N-dimensional space which distinctly classifies the data points (\cite{ekbal2008bengali}). It means, this algorithm clearly draws the decision boundary line between the data points that belong to a particular category and the ones that do not fall into the category. This is applicable to any kind of data that is encoded as a vector. Therefore, if we could produce appropriate vector representations of the data in our hand, we can use SVM to obtain the desired results (\cite{ekbal2008bengali}). Here the input features are the same as in LR that is the Term Frequency Inverse Document Frequency (TF-IDF) values of up to 3 grams. We evaluate the SVM model with L2 regularization.
\subsubsection{Multinomial Naive Bayes (MNB):}
This is a Bayesian classifier that works on the naive assumption of conditional independence of features. This means that each input is independent of the other and this is absolutely unrealistic for real data. Nevertheless, it simplifies several complex tasks and hence validates the need. 

We evaluate a  Naive Bayes classifier for multinomially distributed data, which is derived from Bayes Theorem that finds the probability of a future event given an observed event. MNB is a specialized version of Naive Bayes that is designed more for text documents. Whereas simple naive Bayes would model a document as the presence and absence of particular words, MNB explicitly models the word counts and adjusts the underlying calculations to deal with in. Therefore, the input text data is considered as the bag of words with the count of occurrence of words(frequency) alone considered and the position of words are ignored.

Laplace smoothing is performed using  \(\alpha=1\) to solve the problem of zero probability and then evaluate the MNB model with TF-IDF vectors.
\subsubsection{K-Nearest Neighbour (KNN):}
KNN is used for the classification and regression problems but mostly used for classification task.The KNN algorithm stores all available data and classifies,
on the basis of similarities, a new data point. This implies that it can be conveniently grouped into a
well-suite group using the KNN algorithm as new data emerges. The KNN algorithm assumes that the new upcoming data is related to the available cases and
places the new case into the column that is more similar to the categories available. KNN is a non-parametric algorithm as it does not make any assumption on underlying data  (\cite{nongmeikapam-etal-2017-exploring}). It is often referred to as a lazy learner algorithm because it does not automatically learn from the training set, but instead stores the dataset and performs an operation on the dataset at the time of classification. At the training point, the KNN algorithm only stores the dataset and then classifies the data into a group that is somewhat close to the current data as it encounters new data.

We use KNN for classification with 3,4,5, and 9 neighbours by applying uniform weights.
\subsubsection{Decision Tree (DT):}
The decision tree develops models of classification or
regression in the context of a tree structure.
A dataset is broken down into smaller and smaller subsets, while an associated decision tree is gradually built at the same time. A tree with decision nodes and leaf nodes is the final product. Therefore, a decision tree classification works by generating a tree structure, where each node corresponds to a feature name, and the branches correspond to the feature values. The leaves of the tree represent the classification labels. After sequentially choosing alternative decisions, each node is recursively split again, and finally, the classifier defines some rules to predict the result. Decision trees can accommodate high dimensional data and perform classification without needing much computation. In general, a decision tree classifier has reasonable accuracy.
While speaking about its cons, they are vulnerable to mistakes in classification problems having many classes and a comparatively limited number of training examples. Moreover, it is computationally costly for preparation which implies the method of growing a decision tree is expensive in terms of computation. Each candidate splitting area must be organized at each node before it can find the best split. Combinations of fields are used in some algorithms and a search must be made for optimum combination weights. Pruning algorithms can also be costly, because it is important to shape and compare multiple candidate sub-trees.  Here, maximum depth was 800, and minimum sample splits were 5 for DT. The criteria were Gini and entropy.
\subsubsection{Random Forest (RF):}
Random forest is an ensemble classifier that makes its prediction based on the combination of different decision trees trained on datasets of the same size as training set, called bootstraps, created from a random resampling on the training set itself \citewpar{breiman2001random}. Once a tree is constructed, a set of bootstraps, which do not include any particular record from the original dataset [out-of-bag (OOB) samples], is used as test set. The error rate of the classification of all the test sets is the OOB estimate of the generalization error. RF showed important advantages over other methodologies regarding the ability to handle highly non-linearly correlated data, robustness to noise, tuning simplicity, and opportunity for efficient parallel processing. Moreover, RF presents another important characteristic: an intrinsic feature selection step, applied prior to the classification task, to reduce the variables space by giving an importance value to each feature. RF follows specific rules for tree growing, tree combination, self-testing and post-processing, it is robust to overfitting and it is considered more stable in the presence of outliers and in very high dimensional parameter spaces than other machine learning algorithms \citewpar{caruana2006empirical}. We evaluate the RF model with the same features as DT.
\begin{table*}[!htb] 
\begin{center} 
 \renewcommand{\tabcolsep}{0.8mm}
 \small
\begin{tabular}{|l|r|r|r|r|r|r|r|}
\hline
Classifier & Positive & Negative & Mixed feelings & Neutral state  &  Other language  & Macro Avg & Weighted Avg \\
\hline
Support &2503 &547  &510 &631&211&4402&4402\\
\hline
\multicolumn{8}{|c|}{\textbf{Precision}} \\
\hline
SVM  &0.57 &0.00  &0.00 &0.00&0.00&0.11&0.32\\
MNB &0.59&0.79&0.46&0.50&0.50&0.64&0.59\\
KNN &0.58&0.19&0.13&0.18&0.62&0.34&0.43\\
DT &0.65&0.32&0.23&0.36&0.52&0.42&0.51\\
LR &0.76&0.36&0.24&0.39&0.42&0.36&0.58\\
RF &0.62&0.59&0.71&0.56&0.80&0.66&\textbf{0.63}\\
\hline
\multicolumn{8}{|c|}{\textbf{Recall}} \\
\hline
SVM  &1.00 &0.00  &0.00 &0.00&0.00&0.20&0.57\\
MNB &1.00&0.06&0.00&0.04&0.04&0.28&0.59\\
KNN &0.70&0.04&0.06&0.29&0.07&0.23&0.46\\
DT &0.80&0.23&0.14&0.27&0.37&0.36&0.55\\
LR &0.64&0.43&0.28&0.44&0.64&0.40&0.54\\
RF &0.97&0.17&0.02&0.19&0.43&0.35&\textbf{0.62}\\
\hline
\multicolumn{8}{|c|}{\textbf{F-Score}} \\
\hline
SVM  &0.72 &0.00  &0.00 &0.00&0.00&0.14&0.41\\
MNB & 0.74 & 0.11 & 0.01 & 0.08 & 0.08 &0.28&0.47\\
KNN &0.63&0.07&0.08&0.23&0.13&0.23&0.42\\
DT &0.72&0.27&0.17&0.31&0.44&0.38&0.53\\
LR &0.69&0.39&0.26&0.41&0.51&0.38&\textbf{0.56}\\
RF &0.76&0.26&0.05&0.28&0.56&0.38&0.53\\
\hline
\end{tabular} 
\caption{Precision, Recall, and F-score for Tamil Sentiment Analysis} 
\label{tab:result-tamil-senti} 
\end{center} 
\end{table*}
\begin{table*}[!htb] 
\begin{center} 
 \renewcommand{\tabcolsep}{0.8mm}
 \small
\begin{tabular}{|l|r|r|r|r|r|r|r|}
\hline
Classifier & Positive & Negative & Mixed feelings & Neutral state  &  Other language  & Macro Avg & Weighted Avg \\
\hline
Support &755 &285  &131 &645&146&1962&1962\\
\hline
\multicolumn{8}{|c|}{\textbf{Precision}} \\
\hline
SVM  &0.38 &0.00  &0.00 &0.00&0.00&0.08&0.15\\
MNB &0.49&0.88&0.00&0.60&0.88&0.57&0.58\\
KNN &0.43&0.32&0.41&0.37&0.59&0.42&0.41\\
DT &0.51&0.54&0.35&0.61&0.51&0.50&0.54\\
LR &0.73&0.57&0.34&0.52&0.50&0.53&\textbf{0.59}\\
RF &0.62&0.74&0.56&0.51&0.76&0.64&0.61\\
\hline
\multicolumn{8}{|c|}{\textbf{Recall}} \\
\hline
SVM  &1.00 &0.00 &0.00 &0.00&0.00&0.20&0.38\\
MNB &0.92&0.13&0.00&0.45&0.10&0.32&0.53\\
KNN &0.67&0.12&0.12&0.34&0.21&0.29&0.41\\
DT &0.79&0.32&0.21&0.40&0.42&0.43&0.53\\
LR &0.51&0.45&0.32&0.72&0.66&0.53&0.57\\
RF &0.63&0.31&0.14&0.77&0.41&0.45&\textbf{0.58}\\
\hline
\multicolumn{8}{|c|}{\textbf{F-Score}} \\
\hline
SVM  &0.56 &0.00  &0.00 &0.00&0.00&0.11&0.21\\
MNB &0.64&0.23&0.00&0.52&0.17&0.31&0.46\\
KNN &0.53&0.17&0.19&0.36&0.30&0.31&0.38\\
DT &0.62&0.40&0.26&0.49&0.46&0.44&0.51\\
LR &0.60&0.50&0.33&0.60&0.57&0.52&\textbf{0.57}\\
RF &0.62&0.44&0.22&0.62&0.53&0.49&0.56\\
\hline
\end{tabular} 
\caption{Precision, Recall, and F-score for Malayalam Sentiment Analysis}
\label{tab:result-mal-senti} 
\end{center} 
\end{table*}
\begin{table*}[!htb] 
\begin{center} 
 \renewcommand{\tabcolsep}{0.8mm}
 \small
\begin{tabular}{|l|r|r|r|r|r|r|r|}
\hline
Classifier & Positive & Negative & Mixed feelings & Neutral state  &  Other language  & Macro Avg & Weighted Avg \\
\hline
Support & 363& 162 &57 &83&103&768&768\\
\hline
\multicolumn{8}{|c|}{\textbf{Precision}} \\
\hline
RF & 0.59&0.70 & 0.45 & 0.48 & 0.53 &0.55 &0.58\\
SVM  &0.47 &0.00&0.00  &0.00 &0.00&0.09&0.22\\
MNB &0.54&0.82&1.00&0.75&0.74&0.77&\textbf{0.68}\\
KNN &0.51&0.67&0.44&0.50&0.55&0.53&0.54\\
DT &0.59&0.61&0.21&0.39&0.45&0.45&0.53\\
LR &0.70&0.60&0.24&0.38&0.45&0.47&0.58\\
\hline
\multicolumn{8}{|c|}{\textbf{Recall}} \\
\hline
RF & 0.87&0.48&0.06&0.18&0.50&0.42&\textbf{0.59}\\
SVM  & 1.00&0.00  &0.00 &0.00&0.00&0.20&0.47\\
MNB &0.99&0.36&0.02&0.04&0.14&0.31&0.57\\
KNN &0.91&0.10&0.07&0.05&0.41&0.31&0.52\\
DT &0.73&0.48&0.19&0.14&0.47&0.40&0.54\\
LR &0.69&0.51&0.26&0.36&0.55&0.48&0.57\\
\hline
\multicolumn{8}{|c|}{\textbf{F-Score}} \\
\hline
RF & 0.7&0.57&0.11&0.27&0.52&0.43&0.55\\
SVM  &0.64 &0.00  &0.00 &0.00&0.00&0.13&0.30\\
MNB &0.70&0.50&0.03&0.07&0.23&0.31&0.48\\
KNN &0.65&0.17&0.12&0.09&0.47&0.30&0.43\\
DT &0.66&0.54&0.20&0.21&0.46&0.41&0.52\\
LR &0.70&0.55&0.25&0.37&0.50&0.47&\textbf{0.57}\\
\hline
\end{tabular} 
\caption{Precision, Recall, and F-score for Kannada Sentiment analysis} 
\label{tab:result-kan-senti} 
\end{center} 
\end{table*}
\begin{table*}[!htb] 
\begin{center} 
 \renewcommand{\tabcolsep}{0.8mm}
 \small
\begin{tabular}{|l|r|r|r|r|r|r|r|r|}
\hline
Classifier & Not-O& O-untargeted&OTI &OTG& OT-Other &  Other language  & Macro Avg & Weighted Avg \\
\hline
Support & 3190& 368 & 315 & 288 & 71& 160 &4392&4392\\
\hline
\multicolumn{9}{|c|}{\textbf{Precision}} \\
\hline
RF &0.77 &0.48 &0.65  &0.43  &1.00 &0.88&0.70&0.72 \\ 
SVM &0.73 &0.67 & 0.25& 0.12& 0.00&0.91&0.45&0.65\\
MNB &0.74&0.79&1.00&1.00&0.00&0.96&0.75&\textbf{0.78} \\
KNN &0.73&0.67&0.25&0.12&0.00&0.91&0.45&0.65 \\
DT &0.80&0.29&0.28&0.20&0.11&0.70&0.40&0.67 \\
LR &0.87&0.29&0.27&0.14&0.03&0.68&0.38&0.71 \\
\hline
\multicolumn{9}{|c|}{\textbf{Recall}} \\
\hline
RF &0.99 &0.16 &0.06  &0.03  &0.01 &0.57&0.31&\textbf{0.76} \\ 
SVM&  0.99&0.02&0.01&0.02&0.00&0.13&0.19&0.73\\
MNB &1.00&0.03&0.01&0.00&0.00&0.44&0.25&0.74 \\
KNN &0.99&0.02&0.01&0.02&0.00&0.13&0.19&0.73 \\
DT &0.92&0.20&0.15&0.12&0.03&0.56&0.33&0.72 \\
LR &0.66&0.28&0.30&0.48&0.04&0.72&0.41&0.58 \\
\hline
\multicolumn{9}{|c|}{\textbf{F-Score}} \\
\hline
RF &0.86&0.24&0.12&0.06&0.03&0.69&0.33&\textbf{0.69} \\ 
SVM  & 0.84&0.03&0.01&0.03&0.00&0.23&0.19&0.63\\
MNB &0.85&0.06&0.02&0.01&0.00&0.60&0.26&0.65 \\
KNN &0.84&0.03&0.01&0.03&0.00&0.23&0.19&0.63 \\
DT &0.85&0.24&0.20&0.15&0.04&0.62&0.35&\textbf{0.69} \\
LR &0.75&0.29&0.28&0.22&0.04&0.70&0.38&0.63 \\
\hline
\end{tabular} 
\caption{Precision, Recall, and F-score for Tamil Offensive Language Identification. O-Offensive, T-Targeted, G-Group.} 
\label{tab:result-tamil-off} 
\end{center} 
\end{table*}
\begin{table*}[!htb] 
\begin{center} 
 \renewcommand{\tabcolsep}{0.8mm}
 \small
\begin{tabular}{|l|r|r|r|r|r|r|r|r|}
\hline
Classifier & Not-O& O-untargeted&OTI &OTG& OT-Other &  Other language  & Macro Avg & Weighted Avg \\
\hline
Support &1765& 29 &27 &23&-&157&2001&2001\\
\hline
\multicolumn{9}{|c|}{\textbf{Precision}} \\
\hline
RF &0.95&1.00&1.00&1.00&-&0.95&0.98&\textbf{0.95} \\
SVM &0.88 &0.00 &0.00  &0.00  &- &0.00&0.18&0.78 \\ 
MNB &0.89&0.00&0.00&0.00&-&0.90&0.36&0.86 \\
KNN &0.95&1.00&1.00&1.00&-&0.90&0.97&\textbf{0.95} \\
DT &0.95&0.67&0.79&0.65&-&0.82&0.78&0.93 \\
LR &0.97&0.50&0.33&0.30&-&0.52&0.52&0.91 \\
\hline
\multicolumn{9}{|c|}{\textbf{Recall}} \\
\hline
RF &1.00&0.45&0.37&0.39&-&0.69&0.58&\textbf{0.95} \\
SVM &1.00 &0.00 &0.00  &0.00  &- &0.00&0.20&0.88 \\ 
MNB &1.00&0.00&0.00&0.00&-&0.11&0.22&0.89 \\
KNN &0.99&0.48&0.44&0.43&-&0.68&0.61&\textbf{0.95 }\\
DT &0.98&0.55&0.41&0.48&-&0.69&0.62&0.94 \\
LR &0.89&0.72&0.56&0.52&-&0.85&0.71&0.88 \\
\hline
\multicolumn{9}{|c|}{\textbf{F-Score}} \\
\hline
RF &0.97&0.62&0.54&0.56&-&0.80&0.70&\textbf{0.94} \\
SVM & 0.94&0.00 &0.00  &0.00  &- &0.00&0.19&0.83 \\ 
MNB &0.94&0.00&0.00&0.00&-&0.20&0.23&0.85 \\
KNN &0.97&0.65&0.62&0.61&-&0.78&0.72&\textbf{0.94} \\
DT &0.97&0.60&0.54&0.55&-&0.75&0.68&\textbf{0.94} \\
LR &0.93&0.59&0.42&0.38&-&0.64&0.59&0.89 \\
\hline
\end{tabular} 
\caption{Precision, Recall, and F-score for Malayalam Offensive Language Identification. O-Offensive, T-Targeted, G-Group.} 
\label{tab:result-mal-off} 
\end{center} 
\end{table*}
\begin{table*}[!htb] 
\begin{center} 
 \renewcommand{\tabcolsep}{0.8mm}
 \small
\begin{tabular}{|l|r|r|r|r|r|r|r|r|}
\hline
Classifier & Not-O& O-untargeted&OTI &OTG& OT-Other &  Other language  & Macro Avg & Weighted Avg \\
\hline
Support &427&33  &75 &44&14&185&778&778\\
\hline
\multicolumn{9}{|c|}{\textbf{Precision}} \\
\hline
RF & 0.65 &0.00 &0.71 &0.43 &1.00 &0.67 &0.58&0.63\\
SVM &0.55 &0.00 &0.00  & 0.00 & 0.00&0.00&0.09&0.30 \\ 
MNB &0.60&0.00&0.86&0.00&0.00&0.78&0.37&0.60 \\
KNN &0.61&0.00&0.78&0.67&0.00&0.66&0.45&0.60 \\
DT &0.64&0.21&0.57&0.29&0.25&0.56&0.42&0.57 \\
LR &0.77&0.04&0.63&0.25&0.22&0.64&0.43&\textbf{0.66} \\
\hline
\multicolumn{9}{|c|}{\textbf{Recall}} \\
\hline
RF & 0.89 &0.00 &0.35 & 0.08 &0.06 &0.54 &0.32 &\textbf{0.66}\\
SVM & 1.00&0.00 &0.00  &0.00  &0.00 &0.00&0.17&0.55 \\ 
MNB &0.98&0.00&0.33&0.00&0.00&0.22&0.26&0.62 \\
KNN &0.93&0.00&0.19&0.09&0.00&0.34&0.26&0.61 \\
DT &0.78&0.09&0.51&0.18&0.07&0.45&0.35&0.60 \\
LR &0.76&0.03&0.59&0.23&0.29&0.71&0.43&\textbf{0.66} \\
\hline
\multicolumn{9}{|c|}{\textbf{F-Score}} \\
\hline
RF & 0.75 &0.00 &0.47& 0.14 &0.11 &0.60 & 0.34 &0.61\\
SVM &0.71 &0.00 &0.00  &0.00  &0.00 &0.00&0.12&0.39 \\ 
MNB &0.74&0.00&0.48&0.00&0.00&0.34&0.26&0.54 \\
KNN &0.73&0.00&0.30&0.16&0.00&0.45&0.27&0.55 \\
DT &0.70&0.13&0.54&0.22&0.11&0.50&0.37&0.58 \\
LR &0.77&0.04&0.61&0.24&0.25&0.68&0.43&\textbf{0.66} \\
\hline
\end{tabular} 
\caption{Precision, Recall, and F-score for Kannada Offensive Language Identification. O-Offensive, T-Targeted, G-Group. } 
\label{tab:result-kan-off} 
\end{center} 
\end{table*}
\section{Results and Discussion}
The results of the experiments  with the classifiers described above for both sentiment analysis and offensive language detection are shown in terms of precision, recall, F1-score and support in Table \ref{tab:result-tamil-senti}, Table \ref{tab:result-mal-senti}, Table \ref{tab:result-kan-senti}, Table \ref{tab:result-tamil-off}, Table \ref{tab:result-mal-off}, and Table \ref{tab:result-kan-off}.

We used sklearn\footnote{\url{https://scikit-learn.org/stable/}} to develop the models. A macro-average will compute the metrics (precision, recall, F1-score) independently for each class and average them. Thus this metric treats all classes equally, and it does not take the attribute of class imbalance into account. A weighted average takes the metrics from each class just like a macro average, but the contribution of each class to the average is weighted by the number of examples available for it. The number of comments belonging to different classes from both the tasks are listed as the support values in respective tables.
 
For sentiment analysis, the performance of the various classification algorithms range from being inadequate to average on the code-mixed dataset. Logistic regression, random forest classifiers and decision trees were the ones that fared comparatively better across all sentiment classes. To our surprise, we see that SVM performs poorly, having a worse heterogeneity than the other methods. The precision, recall and F1-score are higher for the ``Positive" class followed by the ``Negative" class. All the other classes performed very poorly. One of the reasons is the nature of the dataset as the classes ``Mixed feelings" and ``Neutral state" are challenging to label for the annotators owing to the problematic examples described before.

For offensive language detection, all the classification algorithms perform equally poorly. We see that logistic regression and random forest are the ones that performed relatively better than the others. The precision, recall and F1-score are higher for the ``Not Offensive" class followed by the ``Offensive Targeted Individual" and ``OL" classes. The reasons for the poor performance of other classes are as same as sentiment analysis. From the tables, we see that the classification algorithms have performed better on the task of sentiment analysis in comparison to that of offensive language detection. One of the main reasons could be the differences in the distributions of the classes among the two different tasks. 

When it comes to sentiment analysis dataset in Kannada, out of the total of 7,671 sentences 46\%  and 19\% belong to the ``Positive" and the ``Negative" classes respectively while the other classes share 9\%,11\% and 15\% respectively. This distribution is better when compared to the Kannada dataset for offensive language detection task where 56\% belong to ``Not Offensive", while the other class share a low distribution of 4\%,8\%,6\%,2\%,24\%. Although the distribution of offensive and non-offensive classes is skewed in all the languages, we were able to observe that overwhelmingly higher percentage of comments belonged to non-offensive class in Tamil and Malayalam datasets than Kannada. 72.4\% of comments in Tamil and 88.44\% comments in Malayalam datasets were non-offensive while in Kannada only 55.79\% of the total comments were non-offensive. This explains why the precision, recall and F-score values of identifying the non-offensive class are consistently higher for Tamil and Malayalam data than Kannada. Next to non-offensive class, the number of comments that belonged to "Not in intended language" class was more than the number of comments belonging to one of the offensive classes in Kannada and Malayalam datasets. In other words, it is easier to recognise the "Not offensive" and "Not in intended language" classes because more comments belong to these two classes than other offensive classes. This trend is shown by the tables \ref{tab:result-tamil-off},  \ref{tab:result-mal-off}, and  \ref{tab:result-kan-off}.

Since we collected the posts from movie trailers, we got more positive sentiment than others as the people who watch trailers are more likely to be interested in movies and this skews the overall distribution. However, as the code-mixing phenomenon is not incorporated in the earlier models, this resource could be taken as a starting point for further research. There is significant room for improvement in code-mixed research with our dataset. In our experiments, we only utilized the machine learning methods, but more information such as linguistic information or hierarchical meta-embedding can be utilized.

\section{Conclusion}
This work introduced code-mixed dataset of the under-resourced Dravidian languages. This data set comprises more than 60,000 comments annotated for sentiment analysis and offensive language identification. To improve the research in the under-resourced Dravidian languages, we created an annotation scheme and achieved a high inter-annotator agreement in terms of Krippendorff $\alpha$ from voluntary annotators on contributions collected using Google Form. We created baselines with gold standard annotated data and presented our results for each class in precision, recall, and F-Score. We expect this resource will enable the researchers to address new and exciting problems in code-mixed research. In future work, we intend to investigate whether we can apply these corpora to build corpora for other under-resourced Dravidian languages.  


\begin{acknowledgements}
Authors Bharathi Raja Chakravarthi, Shardul Suryawanshi, and John Phillip McCrae were supported in part by a research grant from Science Foundation Ireland (SFI) under Grant Number SFI/12/RC/2289$\_$P2 (Insight$\_$2), co-funded by the European Regional Development Fund and Irish Research Council grant IRCLA/2017/129 (CARDAMOM-Comparative Deep Models of Language for Minority and Historical Languages).
\end{acknowledgements}

\bibliographystyle{spbasic}      
\bibliography{ref,anthology}   


\end{document}